\pdfoutput=1

\documentclass[11pt]{article}

\usepackage[preprint]{acl}

\usepackage{times}
\usepackage{latexsym}
\usepackage[T1]{fontenc}

\usepackage[utf8]{inputenc}

\usepackage{microtype}

\usepackage{inconsolata}

\usepackage{graphicx}

\usepackage{times}
\usepackage{bm}
\usepackage{multirow}
\usepackage{xcolor}
\usepackage{colortbl}
\usepackage{algorithm}
\usepackage{algorithmic}
\usepackage{booktabs}
\usepackage{amssymb}
\usepackage{amsmath}
\usepackage{float}

\title{Lifelong Knowledge Editing for LLMs with Retrieval-Augmented Continuous Prompt Learning}

\author{Qizhou Chen$^{1,2}$\thanks{\ \ Q. Chen and T. Zhang contributed equally to this work.}, Taolin Zhang$^{2}$\footnotemark[1], Xiaofeng He$^{1,3}$, Dongyang Li$^{1,2}$, \\
\bf{Chengyu Wang$^{4}$\thanks{\ \ Corresponding author.}, Longtao Huang$^{2}$, Hui Xue$^{2}$} \\
  $^{1}$ East China Normal University, Shanghai, China
  $^{2}$ Alibaba Group, Hangzhou, China\\
  $^{3}$NPPA Key Laboratory of Publishing Integration Development, ECNUP, Shanghai, China\\
  $^{4}$ Alibaba Cloud Computing, Hangzhou, China\\
  \texttt{chen\_qizhou@outlook.com}, \texttt{zhangtl0519@gmail.com},
  \texttt{chengyu.wcy@alibaba-inc.com}
}
\begin{document}
\maketitle

\begin{abstract}
Model editing aims to correct outdated or erroneous knowledge in large language models (LLMs) without the need for costly retraining. Lifelong model editing is the most challenging task that caters to the continuous editing requirements of LLMs. Prior works primarily focus on single or batch editing; nevertheless, these methods fall short in lifelong editing scenarios due to catastrophic knowledge forgetting and the degradation of model performance. Although retrieval-based methods alleviate these issues, they are impeded by slow and cumbersome processes of integrating the retrieved knowledge into the model. In this work, we introduce RECIPE, a RetriEval-augmented ContInuous Prompt lEarning method, to boost editing efficacy and inference efficiency in lifelong learning. RECIPE first converts knowledge statements into short and informative continuous prompts, prefixed to the LLM's input query embedding, to efficiently refine the response grounded on the knowledge. It further integrates the Knowledge Sentinel (KS) that acts as an intermediary to calculate a dynamic threshold, determining whether the retrieval repository contains relevant knowledge. Our retriever and prompt encoder are jointly trained to achieve editing properties, i.e., reliability, generality, and locality. In our experiments, RECIPE is assessed extensively across multiple LLMs and editing datasets, where it achieves superior editing performance.  
RECIPE also demonstrates its capability to maintain the overall performance of LLMs alongside showcasing fast editing and inference speed.
\footnote{Source codes is available at \url{https://github.com/qizhou000/RECIPE}.}

\end{abstract}

\section{Introduction} 

Large language models (LLMs) \cite{DBLP:journals/corr/abs-2302-13971,DBLP:journals/fi/RoumeliotisT23, DBLP:conf/iclr/ZengLDWL0YXZXTM23} have become key techniques in NLP. However, once trained, the knowledge encapsulated within LLMs becomes static \cite{DBLP:conf/emnlp/PetroniRRLBWM19}. This can lead to outputs that are outdated or even erroneous as time progresses \cite{ZJUEditSurvey2023}. In response, model editing techniques have been developed \cite{ROME, MEMIT, GRACE, DBLP:journals/corr/abs-2311-04661, WILKE, LTE}, aimed at efficiently updating and correcting LLMs without the necessity of retraining with large-scale parameters. This concept is economically advantageous as it reduces computational costs and enhances the accuracy of outputs produced by LLMs \cite{KnowledgeEditor, MEND, MEMIT, DBLP:journals/corr/abs-2401-06855}.

\begin{figure*}[!t]
    \centering
    \includegraphics[width=1\textwidth]{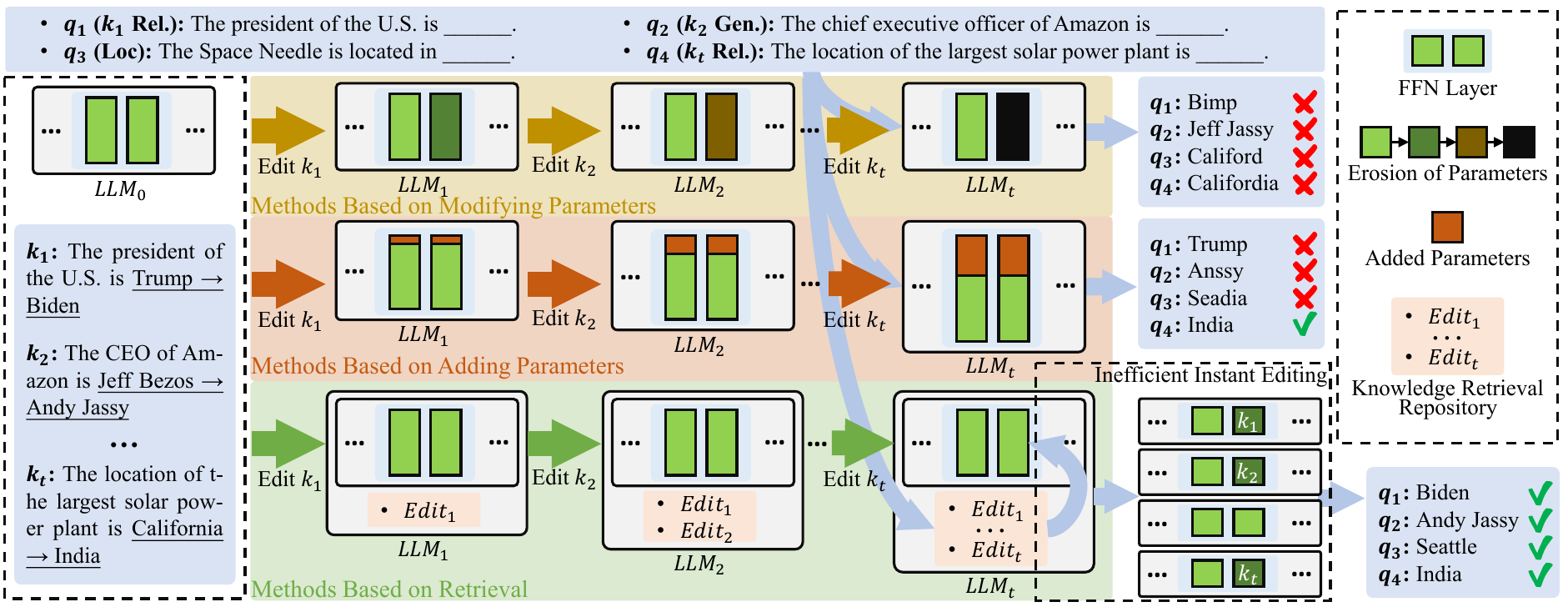}
    \vspace{-0.6cm}
\caption{Comparison among three types of methods in lifelong editing scenarios. 
Modifying parameters and adding extra parameters result in the degradation of LLM performance as editing progresses. In contrast, retrieval-based editors store knowledge in a repository and apply knowledge editing on the fly, which maintains the LLM unchanged and relieves it from accumulating parameter offsets or adding extra parameters. (Best viewed in clolor)} 

    \label{fig_life_long_editing}
\end{figure*}

Previous efforts in model editing have primarily focused on single and batch edits. Notable examples include ROME \cite{ROME}, MEND \cite{MEND}, and MEMIT \cite{MEMIT}, which achieve edits by applying offsets to part of the model's parameters. However, in the real world, LLMs frequently require continuous knowledge updates to stay abreast of emerging knowledge. Thus, the concept of lifelong editing has been introduced \cite{GRACE}. As shown in the upper part of Figure \ref{fig_life_long_editing}, with continuous editing, the accumulating offsets on parameters can result in model performance degradation or even failure \cite{GRACE, T-Patcher, RASE, WILKE}.
Some techniques \cite{CALINET, T-Patcher} address the challenges by integrating extra model parameters. Nevertheless, as shown in the middle of Figure \ref{fig_life_long_editing}, the increase in additional parameters leads to diminished model performance and reduced inference efficiency.

Retrieval-based methods separate knowledge from the model. As shown in the lower part of Figure \ref{fig_life_long_editing}, they temporarily integrate retrieved knowledge into the model during each inference, mitigating knowledge forgetting and performance degradation. However, these methods face certain limitations: retrieval at each token prediction \cite{GRACE}, caching large update matrices \cite{RASE, MELO}, and overly long editing prefixes \cite{LTE}. These shortcomings affect the model's inference efficiency.

In this paper, we introduce RECIPE,
a novel \emph{RetriEval-augmented ContInuous Prompt lEarning} framework to enhance editing efficacy and inference efficiency for LLMs in lifelong learning scenarios. Two key techniques of RECIPE are introduced as follows:

\noindent\textbf{Knowledgeable Continuous Prompt Learning:} 
In retrieval-based methods, LTE \cite{LTE} effectively avoids the shortcomings of earlier approaches. However, their overly long editing prefixes can reduce model inference speed, and full-parameter fine-tuning also increases the risk of overfitting.
Therefore, we aim to explore the shortest possible prefixes to enable the LLM to follow editing instructions. In RECIPE, we achieve this by constructing a continuous short prompt encoder that transforms each piece of editing knowledge (expressed as text) into a knowledgeable continuous prompt.
This approach is grounded in prior research, as exemplified by P-tuning \cite{Prefix-Tuning, p-tuning}, which demonstrated that continuous prompts enable LLMs to perform downstream tasks more effectively. Here, we conceptualize each knowledge edit as a distinct mini-task. To ensure editing efficacy, our prompt encoder is trained to align with three key editing properties including reliability, generality, and locality \cite{ZJUEditSurvey2023}.

\noindent\textbf{Dynamic Prompt Retrieval with Knowledge Sentinel:} 
To avoid unnecessary costs from introducing additional retrievers \cite{RASE,LTE}, we map knowledge statements and queries into the same representational space through the prompt encoder to compute retrieval similarity.
To determine whether the retrieval repository contains knowledge related to an input query, manually setting a fixed similarity threshold is a common practice \cite{RASE}. However, this method does not take into account that different queries often require distinct thresholds due to semantic variations. Therefore, we introduce the Knowledge Sentinel (KS), a trainable embedding representation, as an intermediary to dynamically compute the threshold for each query.
Employing a specifically designed contrastive learning mechanism, the KS module is jointly trained with the prompt encoder to align retrieval with model editing.

The contributions of our paper are summarized as follows:
\begin{itemize}
\item We propose a novel RetriEval-augmented Continuous Prompt Learning framework, RECIPE, to facilitate lifelong model editing. As far as we know, we are the first to adapt prompt learning for model editing, and explore the shortest possible editing prefixes.
\item Within RECIPE, the prompt learning encoder effectively shortens the editing prefixes,  enhancing the inference efficiency of the post-edit LLM. The KS improves editing retrieval, thus enhancing overall editing performance.
\item We conduct comparative experiments of lifelong editing across multiple backbones and editing datasets. The results demonstrate the superiority of RECIPE. 
\end{itemize}

\section{Related Works}
\subsection{Model Editing}
We categorize model editing methods into three types: 
modifying parameters, adding extra parameters, and retrieval-based methods.

\textbf{Methods modifying model parameters} can be further divided into Locate-then-Edit (L\&E) and meta-learning-based methods. For L\&E, ROME \cite{ROME} identifies the LLMs' edit-sensitive layers through causal tracing and proposes rank-one model editing to modify parameters. 
MEMIT \cite{MEMIT} and WILKE \cite{WILKE} respectively use multi-layer allocation and dynamic localization to alleviate the single matrix update burden of ROME.
In meta-learning-based methods, 
KnowledgeEditor \cite{KnowledgeEditor} and MEND \cite{MEND} respectively transform editing knowledge and the gradient decomposition of LLM to the offsets of the weights to be edited. 
MALMEN \cite{DBLP:journals/corr/abs-2311-04661} and DAFNet \cite{DAFNet} enhance MEND's parameter fusion in multiple editing by utilizing normal equations and constructing interactions between edits, respectively.
Although these methods show success in single or batch editing scenarios, in a lifelong editing situation, as the number of edits increases, the accumulating mismatches of parameter offsets can lead to model degradation or failure \cite{WILKE}.

\textbf{Methods adding extra parameters}, such as CaLiNet \cite{CALINET} and T-Patcher \cite{T-Patcher}, achieve model editing by introducing additional neurons to the LLM for each piece of editing knowledge, thereby avoiding modifications to the original model parameters. However, in the lifelong editing scenario, the continuous addition of neurons can progressively dominate the LLM's inference process. This can lead to a reduction in inference speed and model capability.

\textbf{Retrieval-based editors} 
effectively circumvent the issue of accumulated parameter offsets and the potentially unbounded addition of neurons. 
GRACE \cite{GRACE} performs editing retrieval for each token prediction by calculating the linear distance between representations,
which impacts the efficiency of long sequence predictions.
RASE \cite{RASE} develops an editing retrieval model to improve sequential editing. MELO \cite{MELO} introduces a batch editing version of GRACE using LoRA \cite{DBLP:conf/iclr/HuSWALWWC22}. Both methods' instant editing schemes require updating the corresponding FFN matrices separately for each sample in the input batch, which increases memory load as the batch size grows.
LTE \cite{LTE} fine-tunes the LLM to respond to knowledge when prefixed with editing information and retrieves relevant content using the off-the-shelf backbone \cite{DBLP:conf/emnlp/ReimersG19}. However, lengthy editing prefixes still impact inference efficiency. Additionally, fine-tuning the model itself is likely to lead to overfitting, thereby affecting its original performance. 
To address the above issues, I propose RECIPE to explore more efficient retrieval and instant editing, as well as smaller interventions to the model to avoid overfitting on the editing dataset.

\subsection{Prompt Tuning}
Prompt tuning is a typical parameter-efficient learning method that only requires updating a relatively small number of parameters. There are two types of prompt tuning methods: discrete and continuous. The discrete methods \citep{DBLP:conf/acl/GaoFC20,DBLP:conf/acl/LevyBB23, DBLP:conf/acl/WangKMLSKH23, DBLP:conf/nips/DuanDPB23} guide the model to generate relevant outputs for specific tasks by designing fixed, text-based prompts. 
Continuous methods \citep{Prefix-Tuning, p-tuning, DBLP:journals/corr/abs-2110-07602, DBLP:journals/corr/abs-2103-10385, DBLP:conf/nips/Mu0G23,DBLP:conf/wsdm/Xu0QLXHH23,DBLP:conf/acl/ZhangTX00H23}, more relevant to RECIPE, utilize trainable word embedding vectors as prompts. 
Building on the foundations of these works, our approach is duly justified, encoding individual pieces of knowledge as continuous prompts.

\begin{figure*}[!t]
    \centering
    \includegraphics[width=\textwidth]{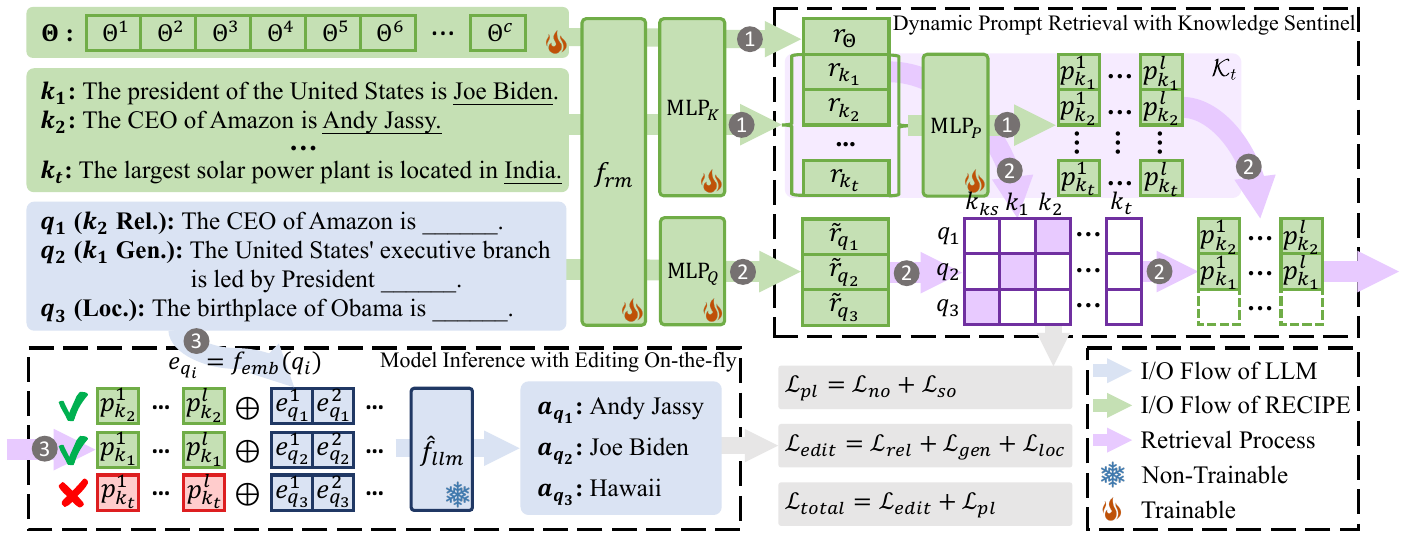}
\caption{Illustration of the RECIPE framework. Process 1 constructs and updates the knowledge retrieval repository $\mathcal{K}_t$. During the inference stage, Process 2 retrieves query-related prompts from $\mathcal{K}_t$. Process 3 utilizes the retrieved continuous prompts to correct the LLM's response. For lifelong editing, the repository can be continuously updated (e.g., from $\mathcal{K}_{t-1}$ to $\mathcal{K}_{t}$) with each new insertion of knowledge and prompts.}
    \label{fig_RECIPE}
\end{figure*}

\section{Background} 
\label{section:background}
In this section, we first formally present the model editing task and its lifelong version. Then, we introduce the evaluation properties in model editing.

An LLM $f_{llm} \in \mathcal{F}$ can be regarded as a function $f_{llm}: \mathcal{Q} \mapsto \mathcal{A}$ that maps an input query $q$ to its predicted answer $a=f_{llm}(q)$.
Given an edit example pair $(q_e, a_e)$ that $f_{llm}(q_e) \neq a_e$, a model editor $\mathbf{ME}: \mathcal{F} \times \mathcal{Q} \times \mathcal{A} \mapsto \mathcal{F}$ outputs a post-edit model $f_{llm}'$ such that:
\begin{equation}
         f_{llm}'=\mathbf{ME}(f_{llm},q_e,a_e)
\end{equation}
Given an initial model $f^{0}_{llm}$, ME will iteratively implement editing as the demands of editing continue to emerge in a lifelong editing scenario:
\begin{gather}
    f^{t}_{llm} = \mathbf{ME}(f^{t-1}_{llm}, q_{e_t},a_{e_t}), t = 1,2,3,...
\end{gather}
At any timestep $t$ in the lifelong editing process, a good ME should make the edited model $f^{t}_{llm}$ meet the following three criteria \cite{ZJUEditSurvey2023}:

\noindent\textbf{Reliability} requires $f^{t}_{llm}$ to correctly remember all the previously edit samples themselves:
\begin{equation}
        \mathbb{E}_{\left(q_e, a_e\right) \sim\left\{\left(q_{e_\tau}, a_{e_\tau}\right)\right\}_{\tau=1}^t} \mathbb{I}\left\{f^{t}_{llm}\left(q_e\right)=a_e\right\}
\end{equation}
where the $\mathbb{I}$ is the indicator function.

\noindent\textbf{Generality} requires $f^{t}_{llm}$ to correctly answer queries belonging to relevant neighbors of previously edited samples:
\begin{equation}
\begin{split}
    \mathbb{E}_{\left(q_e, a_e\right) \sim\left\{\left(q_{e_\tau}, a_{e_\tau}\right)\right\}_{\tau=1}^t}\mathbb{E}_{(q_g, a_g) \sim N\left(q_{e}, a_{e}\right)} \mathbb{I}_g(q_g, a_g)\\
    \text{s.t.}\;\mathbb{I}_g(q_g, a_g) = \mathbb{I}\left\{f^{t}_{llm} \left(q_g\right)=a_g\right\}
\end{split}
\end{equation}
where $N(q_{e}, a_{e})$ is the relevant neighbors of edit sample $(q_{e}, a_{e})$.

\noindent\textbf{Locality} requires $f^{t}_{llm}$ to maintain consistency with the initial model $f^{0}_{llm}$ on queries unrelated to previously edited samples:
\begin{equation}
\begin{split}
    \mathbb{E}_{\left(q_e, a_e\right) \sim\left\{\left(q_{e_\tau}, a_{e_\tau}\right)\right\}_{\tau=1}^t} \mathbb{E}_{(q_l, a_l) \sim O(q_{e}, a_{e})} \mathbb{I}_l(q_l, a_l)\\
    \text{s.t.}\;\mathbb{I}_l(q_l, a_l) = \mathbb{I}\left\{f^{t}_{llm} \left(q_l\right)=f^{0}_{llm}\left(q_l\right)\right\}
\end{split}
\end{equation}
where $O(q_{e}, a_{e})$ is the irrelevant samples set w.r.t. the edit sample $(q_{e}, a_{e})$.
Note that the locality metric implicitly includes the preservation of the general performance of $f^t_{llm}$ relative to $f^0_{llm}$.

\section{The Proposed Approach}

In this section, we formally introduce the RECIPE framework, with the overall architecture in Figure \ref{fig_RECIPE}. First, RECIPE maintains a knowledge retrieval repository, which stores representations of editing knowledge mapped to their knowledgeable continuous prompts described in Sec. \ref{subsection:Updating_KRR}.
Next, we introduce a dynamic retrieval technique with the KS to facilitate knowledge retrieval to filter out irrelevant knowledge in Sec. \ref{subsection:MERKS}.
To ensure the LLMs adhere to the edited knowledge related to the query efficiently, RECIPE prefixes the retrieved continuous prompt to the word embeddings of the LLM's input query, as detailed in Sec. \ref{subsection:GCPE}. Finally, we describe the joint training procedure of the RECIPE framework in Sec. \ref{subsection:train}. 
The algorithms for RECIPE are detailed in Appendix \ref{sec_Algorithms_of_RECIPE}.

\subsection{Construction and Update of Knowledge Retrieval Repository}
\label{subsection:Updating_KRR}
The knowledge retrieval repository is initialized as empty, i.e., $\mathcal{K}_0=\{\}$, and is updated from $\mathcal{K}_{t-1}$ to $\mathcal{K}_t$ by adding a new key-value pair corresponding to new editing knowledge, $k_t$, at each timestep $t$ in our lifelong editing setting. 

Specifically, at timestep $t$, given a new knowledge statement $k_t$, the knowledge representation $r_{k_t}\in \mathcal{R}^{d_r}$ is achieved through an encoder $f_{rm}$ (e.g.,~RoBERTa \cite{roberta}) stacked with a multilayer perceptron (MLP) $\mathbf{MLP}_K$: 
\begin{equation}
    r_{k_t} = \mathbf{MLP}_K(f_{rm}(k_t))
    \label{equation:knowledge_representation}
\end{equation}
where $f_{rm}$ concatenates the maximum, minimum, and average pooling of its output token representations (including the $\mathrm{[CLS]}$ token) into a vector to maximally retain the semantic information of the input. Then, the continuous prompt $p_{k_t}\in \mathbb{R}^{l\times d_{llm}}$ is generated through another MLP, i.e.,~$\mathbf{MLP}_P$:
\begin{equation}
    p_{k_t} = f_{resp}\left(\mathbf{MLP}_P\left(r_{k_t}\right)\right) 
    \label{equation:prompt_embedding}
\end{equation}
where $l$ and $d_{llm}$ are the length of the continuous prompt and the dimension of the LLM's word embedding, respectively. 
In other words, $l$ is the number of Continuous Prompt Tokens (CPTs) leveraged for LLM inference.
$f_{resp}$ is the reshape operation that maps the vector into a matrix with shape $l\times d_{llm}$.
Finally, the knowledge retrieval repository is updated from $\mathcal{K}_{t-1}$ to $\mathcal{K}_{t}$:
$\mathcal{K}_t = \mathcal{K}_{t-1} \cup \{(r_{k_t}, p_{k_t})\}$
where $(r_{k_t}, p_{k_t})$ is the key-value pair for knowledge retrieval.

\subsection{Dynamic Prompt Retrieval with Knowledge Sentinel}
\label{subsection:MERKS}
The existence of a query-related prompt in the repository is usually determined by using a manually set similarity threshold \cite{RASE}. However, using a fixed threshold does not account for the fact that the sensitivity to similarity with related knowledge varies among different queries due to semantic differences.
The Knowledge Sentinel (KS) serves as an intermediary leveraged to dynamically compute similarity thresholds for various queries. To be specific, KS $\Theta\in\mathcal{R}$ is a trainable word embedding of $f_{rm}$ with token length $c$. It is transformed into the knowledge representation space as:  
$r_{\Theta}= \mathbf{MLP}_{K}(f_{rm}(\Theta))$.
Given a query $q$ and the knowledge retrieval repository $\mathcal{K}_t=\{(r_{k_\tau}, p_{k_\tau})\}_{\tau=1}^t$, the prompt retrieval process is as follows:
\begin{equation}
    \label{equation:query_representation}
    \tilde{r}_{q} = \mathbf{MLP}_Q(f_{rm}(q))
\end{equation}
\begin{equation}
\begin{split}
\mathbf{KS}(q) = 
\begin{cases}
        p_{k_j} & \tilde{r}_{q}^T \cdot r_{k_{j}} > \tilde{r}_{q}^T \cdot r_{\Theta}\\
        \emptyset & \text{otherwise}
\end{cases}
\end{split}
\end{equation}
where $j = {\mathrm{argmax}}_{\tau=1,\ldots,t}\ \tilde{r}_{q}^T \cdot r_{k_\tau}$,
which can be efficiently searched via modern vector databases or search engines (e.g.,~\citet{DBLP:conf/nips/ChenZWLLLYW21}). $\mathbf{MLP}_Q$ is the MLP that maps the query representation to the knowledge representation space. If the retrieved continuous prompt is not sufficiently similar to the query compared to KS, an empty set is returned. Hence, the inference of LLMs is not modified.

\subsection{Model Inference with Editing On-the-fly}
\label{subsection:GCPE}
Previous retrieval-based methods suffer from cumbersome editing processes and post-retrieval knowledge integration \cite{GRACE, LTE}. To address this challenge, we prefix the retrieved continuous prompt to the word embedding of the input query to efficiently correct the response of the LLM.

Specifically, we consider the LLM to be edited as $f_{llm}:\mathcal{Q}\mapsto \mathcal{A}$, where $\hat{f}_{llm}$ is $f_{llm}$ with the embedding layer $f_{emb}$ removed. Given an input query $q$, and the retrieved continuous prompt $p_{k_\tau} = \mathbf{KS}(q)$, the inference process is reformulated as:
$a_{q} = \hat{f}_{llm}(p_{k_\tau} \oplus f_{emb}(q))$
where $\oplus$ denotes the concatenation of the retrieved continuous prompt matrix and the word embedding matrix of $q$. 

The feasibility of our approach is supported by previous work such as P-Tuning \cite{Prefix-Tuning, p-tuning}, which demonstrates the efficacy of training continuous prompt embeddings to enhance the performance of LLMs on downstream tasks. In RECIPE, we treat the editing of each knowledge statement as a mini-task. Instead of fine-tuning a specific prompt encoder for each mini-task, we accomplish the objectives of these mini-tasks by training RECIPE modules that generate continuous prompts, ensuring the LLM adheres to the corresponding knowledge.

\subsection{Model Training}
\label{subsection:train} 
The losses are formulated to ensure adherence to the editing of generated continuous prompts and effective retrieval of query-related knowledge for the LLM.
Given a batch of training data consisting of
$b$ editing sample pairs $\{(q_{e_i}, a_{e_i})\}_{i=1}^b$ 
and their corresponding sampled generality and locality pairs
$\{(q_{g_i},a_{g_i})\}_{i=1}^{b}, \{(q_{l_i},a_{l_i})\}_{i=1}^{b}$, 
the losses are formulated as follows.

\noindent\textbf{Editing:} The editing loss aims to ensure that the generated continuous prompt guides the LLM to follow the properties of reliability, generality, and locality \cite{ZJUEditSurvey2023}. 
Based on the pairs $(q_{e_i}, a_{e_i})$, the sample-wise losses corresponding to these three properties are defined as follows:
\begin{gather}
    \mathcal{L}^{(i)}_{rel} = -\log \hat{f}_{llm}\left(a_{e_i}\left|p_{k_i}\oplus f_{emb}(q_{e_i})\right.\right)\label{eq_editing_loss_rel}\\
    \mathcal{L}^{(i)}_{gen} = -\log \hat{f}_{llm}\left(a_{g_i}\left|p_{k_i}\oplus f_{emb}(q_{g_i})\right.\right)\label{eq_editing_loss_gen}\\
    \mathcal{L}^{(i)}_{loc} =  \mathrm{KL}\left(f_{llm}\left(q_{l_i}\right) || \hat{f}_{llm}\left(p_{k_i}\oplus f_{emb}(q_{l_i})\right)\right)
    \label{eq_editing_loss_loc}
\end{gather}
where $p_{k_i}$ is the continuous prompt transformed through Eq. \ref{equation:knowledge_representation} and Eq. \ref{equation:prompt_embedding} using knowledge $k_i$ that is the concatenation of $q_{e_i}$ and $a_{e_i}$. 
The $\mathrm{KL}$ denotes the Kullback-Leibler divergence, 
which is chosen to better fit the LLM's original prediction distribution on the locality data.
The batch-wise loss function for model editing is derived as follows: 
\begin{equation}
    \mathcal{L}_{edit} = \frac{1}{b}\sum_{i=1}^b \left( \mathcal{L}^{(i)}_{rel} + \mathcal{L}^{(i)}_{gen} + \mathcal{L}^{(i)}_{loc} \right).
    \label{eq_loss_edit}
\end{equation}

\noindent\textbf{Prompt Learning:}
The training losses for prompt learning are based on contrastive learning \cite{DBLP:journals/corr/abs-1807-03748, DBLP:conf/cvpr/He0WXG20} and are aligned with the properties of reliability, generality, and locality \cite{ZJUEditSurvey2023}. For a batch of samples, the loss functions for learning continuous prompts are formulated as follows:
\begin{equation}
\mathcal{L}^{(i)}_{no}  = \delta(\tilde{r}_{q_{e_i}}, r_{k_i}, R) 
    + \delta(\tilde{r}_{q_{g_i}}, r_{k_i}, R),
    \label{eq_retrieval_loss_no}
\end{equation}
\begin{equation}
\begin{split}
\mathcal{L}^{(i)}_{so} = & \delta(\tilde{r}_{q_{l_i}}, r_{\Theta},R)
    + \delta(\tilde{r}_{q_{e_i}},r_{\Theta}, R_{\backslash k_i})\\
&    + \delta(\tilde{r}_{q_{g_i}}, r_{\Theta}, R_{\backslash k_i}),    
    \label{eq_retrieval_loss_so}
\end{split}
\end{equation}
\begin{equation}
\mathcal{L}_{pl} = \frac{1}{b}\sum_{i=1}^b (\mathcal{L}^{(i)}_{no} + \mathcal{L}^{(i)}_{so}),
    \label{eq_loss_pl}
\end{equation}
where $R = \{r_{k_i}\}_{i=1}^b \cup \{r_{\Theta}\}$ and $R_{\backslash k_i} = R \setminus \{r_{k_i}\}$. $r_{k_i}$ is the representation of the editing knowledge $k_i$ transformed through Eq. \ref{equation:knowledge_representation}. The query representations $\tilde{r}_{q_{e_i}}, \tilde{r}_{q_{g_i}}, \tilde{r}_{q_{l_i}}$ for $q_{e_i}, q_{g_i}, q_{l_i}$ are attained via Eq. \ref{equation:query_representation}, respectively. $\delta$ is the InfoNCE loss \cite{DBLP:journals/corr/abs-1807-03748}, formulated as:
\begin{equation}
    \delta(q, k_+,\{k_i\}_{i=1}^n) = -\log \frac{\exp(q \cdot k_+ / \tau)}{\sum_{i=1}^n \exp(q \cdot k_i / \tau)},
\end{equation}
where $\tau$ is the temperature, typically set to 1 by default.
In our work, the \emph{neighbor-oriented} loss $\mathcal{L}^{(i)}_{no}$ encourages higher similarity between the editing knowledge and the corresponding reliability or generality queries. The \emph{sentinel-oriented} loss $\mathcal{L}^{(i)}_{so}$ ensures that input queries yield the highest similarity with the KS in cases where the retrieval repository lacks relevant knowledge.

Thus, the total training loss is:
$\mathcal{L}_{total} = \mathcal{L}_{edit} + \mathcal{L}_{pl}$.
During training, the parameters of the LLM $f_{llm}$ are kept frozen. The trainable modules include only $f_{rm}$, $\mathbf{MLP}_K$, $\mathbf{MLP}_Q$, $\mathbf{MLP}_P$, and $\Theta$, which renders our approach highly lightweight.

\begin{table*}[!tb]
    \scriptsize
    \centering
    \setlength{\tabcolsep}{3.8pt}
    \renewcommand{\arraystretch}{0.85}
    \begin{tabular}{cccccccccccccccc}

\toprule
 &\multirow{2}{*}{\textbf{\# Editing}} & \multirow{2}{*}{\textbf{Type}} & \multirow{2}{*}{\textbf{Editor}} & \multicolumn{4}{c}{\textbf{ZSRE}}                                          & \multicolumn{4}{c}{\textbf{CF}}                                            & \multicolumn{4}{c}{\textbf{RIPE}}                                          \\
 &                &            &                         & \textbf{Rel.}           & \textbf{Gen.}           & \textbf{Loc.}           & \textbf{Avg.}           & \textbf{Rel.}           & \textbf{Gen.}           & \textbf{Loc.}           & \textbf{Avg.}           & \textbf{Rel.}           & \textbf{Gen.}           & \textbf{Loc.}           & \textbf{Avg.}           \\ 
\midrule
&\multirow{11}{*}{1}&  \multirow{6}{*}{MP}&FT &47.86 &42.57 &93.89 &61.44$_{(\pm1.00)}$ &41.37 &26.04 &52.25 &39.89$_{(\pm0.74)}$ &41.54 &33.89 &53.27 &42.90$_{(\pm0.33)}$ \\
&& &MEND &73.86 &70.33 &66.10 &70.10$_{(\pm0.96)}$ &81.06 &67.15 &77.13 &75.11$_{(\pm0.62)}$ &66.37 &29.37 &29.68 &41.81$_{(\pm0.91)}$ \\
&& &ROME &53.49 &51.58 &93.96 &66.34$_{(\pm0.69)}$ &41.07 &21.82 &91.85 &51.58$_{(\pm0.82)}$ &48.33 &27.08 &42.48 &39.30$_{(\pm0.89)}$ \\
&& &MEMIT &49.67 &49.36 &91.87 &63.64$_{(\pm0.61)}$ &45.40 &29.25 &92.93 &55.86$_{(\pm0.39)}$ &58.37 &29.54 &38.67 &42.19$_{(\pm0.39)}$ \\
&& &MALMEN &46.37 &47.75 &33.73 &42.62$_{(\pm0.43)}$ &52.45 &42.31 &36.58 &43.78$_{(\pm0.58)}$ &51.53 &33.86 &20.45 &35.28$_{(\pm1.05)}$ \\
&& &WILKE &50.71 &48.52 &93.31 &64.18$_{(\pm0.55)}$ &40.07 &21.92 &91.70 &51.23$_{(\pm0.45)}$ &47.85 &27.90 &38.50 &38.08$_{(\pm1.02)}$ \\
\cmidrule(lr){3-16}
&& \multirow{1}{*}{AP}&TP &86.35 &83.98 &86.34 &85.56$_{(\pm0.53)}$ &91.41 &68.61 &38.94 &66.32$_{(\pm1.18)}$ &76.98 &55.10 &51.29 &61.13$_{(\pm0.48)}$ \\
\cmidrule(lr){3-16}
&& \multirow{4}{*}{RB}&GRACE &99.20 &33.23 &99.82 &77.42$_{(\pm0.78)}$ &98.65 &11.42 &98.73 &69.60$_{(\pm0.66)}$ &98.13 &28.45 &99.75 &75.44$_{(\pm0.65)}$ \\
&& &R-ROME &51.87 &49.40 &98.82 &66.70$_{(\pm1.54)}$ &39.46 &20.76 &97.38 &52.54$_{(\pm0.86)}$ &46.15 &23.95 &92.99 &54.37$_{(\pm0.96)}$ \\
&& &LTE &98.97 &97.29 &85.90 &94.05$_{(\pm0.15)}$ &98.12 &97.13 &92.20 &95.81$_{(\pm1.21)}$ &98.49 &88.09 &85.79 &90.79$_{(\pm0.61)}$ \\
&& &RECIPE &\textbf{99.40} &\textbf{99.01} &\textbf{99.96} &\textbf{99.46}$_{(\pm0.07)}$ &\textbf{98.78} &\textbf{98.78} &\textbf{99.01} &\textbf{98.86}$_{(\pm0.39)}$ &\textbf{99.36} &\textbf{89.56} &\textbf{99.78} &\textbf{96.24}$_{(\pm0.95)}$ \\
\midrule
&\multirow{11}{*}{10}& \multirow{6}{*}{MP}&FT &44.08 &43.98 &70.03 &52.70$_{(\pm0.30)}$ &18.09 &15.49 &21.53 &18.37$_{(\pm1.39)}$ &22.74 &18.51 &18.84 &20.03$_{(\pm0.53)}$ \\
&& &MEND &0.31 &0.30 &3.31 &1.31$_{(\pm0.29)}$ &0.01 &0.01 &0.07 &0.03$_{(\pm0.01)}$ &0.30 &0.24 &1.64 &0.73$_{(\pm0.14)}$ \\
&& &ROME &41.08 &39.62 &93.02 &57.91$_{(\pm0.69)}$ &38.59 &24.95 &83.60 &49.05$_{(\pm0.53)}$ &33.38 &20.26 &29.53 &27.72$_{(\pm0.50)}$ \\
&& &MEMIT &24.28 &24.14 &51.12 &33.18$_{(\pm0.42)}$ &18.66 &15.39 &62.89 &32.31$_{(\pm1.26)}$ &18.42 &13.63 &10.13 &14.06$_{(\pm1.04)}$ \\
&& &MALMEN &96.22 &88.32 &92.57 &92.37$_{(\pm1.19)}$ &79.52 &45.84 &56.18 &60.52$_{(\pm1.05)}$ &84.75 &47.50 &70.88 &67.71$_{(\pm1.10)}$ \\
&& &WILKE &46.00 &43.02 &86.88 &58.64$_{(\pm0.76)}$ &39.51 &20.35 &86.01 &48.62$_{(\pm0.63)}$ &40.34 &24.73 &27.39 &30.82$_{(\pm0.22)}$ \\
\cmidrule(lr){3-16}
&& \multirow{1}{*}{AP}&TP &57.33 &52.37 &36.68 &48.79$_{(\pm1.47)}$ &85.92 &58.64 &21.56 &55.37$_{(\pm0.43)}$ &63.38 &41.20 &30.45 &45.01$_{(\pm0.80)}$ \\
\cmidrule(lr){3-16}
&& \multirow{4}{*}{RB}&GRACE &52.10 &36.64 &98.80 &62.51$_{(\pm0.78)}$ &60.61 &11.89 &96.52 &56.34$_{(\pm0.78)}$ &49.14 &30.68 &98.30 &59.37$_{(\pm0.96)}$ \\
&& &R-ROME &51.03 &46.40 &97.45 &64.96$_{(\pm0.49)}$ &38.82 &19.50 &95.17 &51.16$_{(\pm1.26)}$ &45.54 &22.53 &85.45 &51.17$_{(\pm1.40)}$ \\
&& &LTE &97.21 &97.18 &85.02 &93.14$_{(\pm0.59)}$ &97.91 &97.01 &92.87 &95.93$_{(\pm1.03)}$ &98.14 &87.40 &85.54 &90.36$_{(\pm0.13)}$ \\
&& &RECIPE &\textbf{99.11} &\textbf{98.82} &\textbf{99.98} &\textbf{99.31}$_{(\pm0.43)}$ &\textbf{98.40} &\textbf{99.11} &\textbf{98.70} &\textbf{98.74}$_{(\pm0.38)}$ &\textbf{98.43} &\textbf{87.98} &\textbf{99.02} &\textbf{95.14}$_{(\pm0.45)}$ \\
\midrule
&\multirow{11}{*}{100}& \multirow{6}{*}{MP}&FT &35.95 &33.35 &25.30 &31.54$_{(\pm1.06)}$ &3.63 &0.19 &0.01 &1.27$_{(\pm0.32)}$ &9.36 &5.19 &6.19 &6.91$_{(\pm1.00)}$ \\
&& &MEND &0.01 &0.03 &0.10 &0.04$_{(\pm0.01)}$ &0.03 &0.01 &0.16 &0.06$_{(\pm0.01)}$ &0.02 &0.01 &0.01 &0.01$_{(\pm0.00)}$ \\
&& &ROME &9.54 &10.43 &21.99 &13.99$_{(\pm0.33)}$ &33.61 &22.09 &68.06 &41.25$_{(\pm1.54)}$ &5.90 &4.16 &5.20 &5.09$_{(\pm1.18)}$ \\
&& &MEMIT &0.76 &0.72 &0.86 &0.78$_{(\pm0.46)}$ &0.63 &0.59 &3.68 &1.63$_{(\pm0.27)}$ &0.20 &0.45 &0.30 &0.32$_{(\pm0.06)}$ \\
&& &MALMEN &54.28 &51.77 &65.25 &57.10$_{(\pm0.88)}$ &48.07 &22.43 &47.20 &39.23$_{(\pm0.75)}$ &66.59 &45.71 &58.54 &56.95$_{(\pm1.06)}$ \\
&& &WILKE &21.48 &20.33 &42.67 &28.16$_{(\pm0.16)}$ &34.39 &19.38 &75.34 &43.03$_{(\pm0.81)}$ &27.91 &17.23 &25.73 &23.62$_{(\pm0.76)}$ \\
\cmidrule(lr){3-16}
&& \multirow{1}{*}{AP}&TP &46.05 &41.20 &9.67 &32.30$_{(\pm0.91)}$ &70.01 &40.76 &4.51 &38.42$_{(\pm0.54)}$ &44.73 &28.94 &11.60 &28.42$_{(\pm0.91)}$ \\
\cmidrule(lr){3-16}
&& \multirow{4}{*}{RB}&GRACE &47.62 &34.99 &98.00 &60.21$_{(\pm0.41)}$ &55.00 &12.85 &93.49 &53.78$_{(\pm0.54)}$ &41.03 &31.02 &95.15 &55.74$_{(\pm0.44)}$ \\
&& &R-ROME &50.50 &41.70 &96.02 &62.74$_{(\pm0.56)}$ &36.99 &17.06 &92.76 &48.94$_{(\pm0.85)}$ &44.76 &19.52 &77.03 &47.10$_{(\pm0.43)}$ \\
&& &LTE &95.18 &93.39 &85.11 &91.23$_{(\pm0.69)}$ &96.28 &96.01 &91.94 &94.74$_{(\pm1.19)}$ &96.87 &85.59 &84.73 &89.06$_{(\pm1.51)}$ \\
&& &RECIPE &\textbf{97.78} &\textbf{97.04} &\textbf{99.98} &\textbf{98.27}$_{(\pm0.15)}$ &\textbf{96.68} &\textbf{97.05} &\textbf{96.53} &\textbf{96.75}$_{(\pm1.06)}$ &\textbf{97.48} &\textbf{87.21} &\textbf{95.60} &\textbf{93.43}$_{(\pm0.31)}$ \\
\midrule
&\multirow{11}{*}{1000}& \multirow{6}{*}{MP}&FT &14.66 &12.61 &2.69 &9.99$_{(\pm1.00)}$ &6.94 &0.68 &3.48 &3.70$_{(\pm0.09)}$ &7.91 &2.13 &1.82 &3.95$_{(\pm0.40)}$ \\
&& &MEND &0.04 &0.02 &0.00 &0.02$_{(\pm0.01)}$ &0.01 &0.00 &0.02 &0.01$_{(\pm0.00)}$ &0.00 &0.02 &0.02 &0.02$_{(\pm0.00)}$ \\
&& &ROME &1.54 &1.48 &0.63 &1.22$_{(\pm0.90)}$ &0.15 &0.13 &0.12 &0.14$_{(\pm0.03)}$ &0.02 &0.01 &0.03 &0.02$_{(\pm0.01)}$ \\
&& &MEMIT &0.18 &0.22 &0.14 &0.18$_{(\pm0.07)}$ &0.09 &0.05 &0.99 &0.38$_{(\pm0.18)}$ &0.02 &0.02 &0.03 &0.02$_{(\pm0.01)}$ \\
&& &MALMEN &32.03 &28.50 &28.14 &29.56$_{(\pm1.33)}$ &15.80 &16.41 &22.53 &18.25$_{(\pm0.22)}$ &42.33 &38.45 &38.52 &39.77$_{(\pm0.97)}$ \\
&& &WILKE &15.19 &12.60 &25.31 &17.70$_{(\pm1.32)}$ &13.22 &12.28 &43.09 &22.86$_{(\pm0.64)}$ &15.19 &14.25 &10.99 &13.48$_{(\pm1.15)}$ \\
\cmidrule(lr){3-16}
&& \multirow{1}{*}{AP}&TP &44.72 &41.38 &4.38 &30.16$_{(\pm1.04)}$ &64.70 &32.50 &11.63 &36.28$_{(\pm0.72)}$ &42.24 &26.80 &9.87 &26.30$_{(\pm1.01)}$ \\
\cmidrule(lr){3-16}
&& \multirow{4}{*}{RB}&GRACE &42.04 &33.42 &96.73 &57.40$_{(\pm0.68)}$ &52.75 &12.86 &91.02 &52.21$_{(\pm0.85)}$ &38.03 &30.10 &91.24 &53.12$_{(\pm0.61)}$ \\
&& &R-ROME &48.73 &36.49 &94.09 &59.77$_{(\pm0.77)}$ &35.64 &14.03 &87.94 &45.87$_{(\pm0.91)}$ &41.49 &16.96 &68.98 &42.48$_{(\pm1.21)}$ \\
&& &LTE &93.03 &91.14 &84.42 &89.53$_{(\pm1.16)}$ &95.87 &95.27 &89.35 &93.50$_{(\pm0.26)}$ &94.53 &84.52 &80.44 &86.50$_{(\pm0.75)}$ \\
&& &RECIPE &\textbf{96.30} &\textbf{95.27} &\textbf{99.98} &\textbf{97.18}$_{(\pm0.50)}$ &\textbf{96.37} &\textbf{96.04} &\textbf{93.66} &\textbf{95.35}$_{(\pm0.61)}$ &\textbf{95.60} &\textbf{85.53} &\textbf{92.35} &\textbf{91.16}$_{(\pm1.28)}$ \\
\midrule
&\multirow{11}{*}{10000}& \multirow{6}{*}{MP}&FT &6.57 &5.29 &0.44 &4.10$_{(\pm0.24)}$ &4.86 &0.76 &2.19 &2.60$_{(\pm1.40)}$ &- &- &- &- \\
&& &MEND &0.01 &0.02 &0.01 &0.01$_{(\pm0.01)}$ &0.03 &0.01 &0.00 &0.01$_{(\pm0.00)}$ &- &- &- &- \\
&& &ROME &1.21 &0.38 &0.15 &0.58$_{(\pm0.19)}$ &0.16 &0.07 &0.22 &0.15$_{(\pm0.07)}$ &- &- &- &- \\
&& &MEMIT &0.03 &0.01 &0.02 &0.02$_{(\pm0.00)}$ &0.03 &0.02 &0.03 &0.02$_{(\pm0.01)}$ &- &- &- &- \\
&& &MALMEN &15.75 &10.82 &17.99 &14.85$_{(\pm0.39)}$ &6.14 &5.50 &8.17 &6.60$_{(\pm1.01)}$ &- &- &- &- \\
&& &WILKE &7.39 &5.11 &14.05 &8.85$_{(\pm1.15)}$ &5.17 &3.70 &23.87 &10.91$_{(\pm1.17)}$ &- &- &- &- \\
\cmidrule(lr){3-16}
&& \multirow{1}{*}{AP}&TP &37.53 &33.55 &3.94 &25.01$_{(\pm0.56)}$ &58.26 &29.25 &11.42 &32.98$_{(\pm0.67)}$ &- &- &- &- \\
\cmidrule(lr){3-16}
&& \multirow{4}{*}{RB}&GRACE &38.50 &31.52 &93.15 &54.39$_{(\pm0.43)}$ &48.52 &11.75 &85.38 &48.55$_{(\pm1.78)}$ &- &- &- &- \\
&& &R-ROME &45.94 &27.04 &91.20 &54.73$_{(\pm0.75)}$ &33.72 &10.42 &84.16 &42.77$_{(\pm0.63)}$ &- &- &- &- \\
&& &LTE &88.80 &86.94 &83.38 &86.37$_{(\pm0.88)}$ &93.10 &91.55 &84.32 &89.66$_{(\pm0.87)}$ &- &- &- &- \\
&& &RECIPE &\textbf{93.79} &\textbf{91.32} &\textbf{99.64} &\textbf{94.92}$_{(\pm0.70)}$ &\textbf{95.51} &\textbf{93.76} &\textbf{90.82} &\textbf{93.36}$_{(\pm1.68)}$ &- &- &- &- \\
\bottomrule
    \end{tabular} 
    \caption{The overall results using LLAMA-2 (7B) in lifelong editing scenario. Editing results of GPT-J and GPT-XL are shown in Appendix \ref{result_diff_llm}. ``\# Editing'' denotes the number of edits. ``Rel.'', ``Gen.'' and ``Loc.'' are the abbreviations of reliability, generality, and locality, respectively. Given that the RIPE dataset comprises 4,388 samples, achieving results for 10,000 edits is not feasible. MP, AP, and RB indicate Modifying Parameters, Adding Parameters, and Retrieval-Based methods, respectively. The t-tests demonstrate the improvements of our work are statistically significant with $p$ < 0.05 level.}
    \label{tab_main_exp}
\end{table*}

\begin{table}[!tb]
    \footnotesize
    \renewcommand{\arraystretch}{0.9}
    \centering
    \setlength{\tabcolsep}{2.6pt}

\begin{tabular}{lccccc}
        \toprule
\textbf{Editor} & \textbf{CSQA}  & \textbf{MMLU}  & \textbf{ANLI}  & \textbf{SQUAD-2} & \textbf{Average} \\
        \midrule
N/A             & 38.91          & 41.54          & 34.04          & 36.43            & 37.73            \\
        \midrule
FT              & 19.27          & 29.93          & 33.33          & 0.59             & 20.78            \\
MEND            & 20.31          & 24.68          & 33.07          & 0.04             & 19.52            \\
ROME            & 19.97          & 23.03          & 33.47          & 0.41             & 19.22            \\
MEMIT           & 19.68          & 23.23          & 33.39          & 0.01             & 19.08            \\
TP              & 19.62          & 22.84          & 33.37          & 0.96             & 19.20            \\
GRACE           & 38.60          & 41.20          & 33.93          & 36.28            & 37.50            \\
R-ROME          & 38.50          & 41.12          & 33.90          & 36.31            & 37.46            \\
MALMEN          & 20.85          & 24.83          & 33.03          & 0.27             & 19.75            \\
LTE             & 19.45          & 23.21          & 33.41          & 25.25            & 25.33            \\
WILKE           & 19.87          & 23.37          & 33.37          & 0.07             & 19.17            \\
RECIPE          & \textbf{38.76} & \textbf{41.40} & \textbf{34.13} & \textbf{36.50}   & \textbf{37.70}  \\
        \bottomrule
\end{tabular}
 
    \caption{Performance of LLAMA-2 after 1,000 edits. ``N/A'' denotes performance without any edits. Bold font highlights the optimal post-editing performance.}
    \label{tab_llm_gen_exp}
\end{table}

\begin{table}[!tb]
    \footnotesize
    \centering
    \setlength{\tabcolsep}{3.6pt}

\begin{tabular}{ccccc}
        \toprule
\textbf{Type}                         & \textbf{Editor} & \textbf{Edit Time}                      & \textbf{Infer. Time}                  & \textbf{Total Time}                      \\
        \midrule
                             & FT     & 1.7205                         & \cellcolor[HTML]{9BC2E6}0.0589 & 1.7794                         \\
                             & MEND   & \cellcolor[HTML]{DDEBF7}0.0987 & \cellcolor[HTML]{9BC2E6}0.0590 & 0.1577                         \\
                             & ROME   & 17.1639                        & \cellcolor[HTML]{9BC2E6}0.0586 & 17.2225                        \\
                             & MEMIT  & 33.6631                        & \cellcolor[HTML]{9BC2E6}0.0591 & 33.7222                        \\
                             & MALMEN & 2.3418                         & \cellcolor[HTML]{9BC2E6}0.0589 & 2.4007                         \\
\multirow{-6}{*}{MP} & WILKE  & 38.7165                        & \cellcolor[HTML]{9BC2E6}0.0587 & 38.7752                        \\
        \midrule
AP                    & TP     & 5.9061                         & 0.0615                         & 5.9676                         \\
        \midrule
                             & GRACE  & 12.5343                        & 0.0936                         & 12.6279                        \\
                             & R-ROME & 17.3135                        & 0.0606                         & 17.3741                        \\
                             & LTE    & \cellcolor[HTML]{9BC2E6}0.0076 & 0.0634                         & \cellcolor[HTML]{DDEBF7}0.0710 \\
\multirow{-4}{*}{RB}  & RECIPE   & \cellcolor[HTML]{9BC2E6}0.0078 & \cellcolor[HTML]{DDEBF7}0.0598 & \cellcolor[HTML]{9BC2E6}0.0676\\
        \bottomrule
\end{tabular}

    \caption{Average time (s) taken for a single edit and model inference after 10,000 edits. }
    \label{tab_time_compare}
\end{table}

\section{Experiments}
In this section, we present the experimental results on LLMs including LLAMA-2 (7B), GPT-J (6B), and GPT-XL (1.5B). We compare RECIPE against strong baselines including Fine-Tune (FT) \cite{ZJUEditSurvey2023}, MEND \cite{MEND}, ROME \cite{ROME}, MEMIT \cite{MEMIT}, MALMEN \cite{DBLP:journals/corr/abs-2311-04661}, WILKE \cite{WILKE}, T-Patcher (TP) \cite{T-Patcher}, GRACE \cite{GRACE}, R-ROME \cite{RASE}, and LTE \cite{LTE}. For the evaluation of lifelong editing, we use datasets including ZSRE \cite{MEND}, CF \cite{ROME}, and RIPE \cite{RIPPLEDITS}. For the evaluation of general performance after editing, we apply datasets including CSQA \cite{DBLP:conf/naacl/TalmorHLB19}, ANLI \cite{DBLP:conf/acl/NieWDBWK20}, MMLU \cite{DBLP:conf/iclr/HendrycksBBZMSS21}, and SQuAD-2 \cite{DBLP:conf/acl/RajpurkarJL18}. For more detailed description of datasets, baselines, and model settings, please refer to Appendix \ref{datasets_baselines} and Appendix \ref{model_settings}.


\subsection{The Performance of RECIPE}

\noindent\textbf{Editing Performance:} Table \ref{tab_main_exp} and Appendix \ref{result_diff_llm} present the overall editing performance across various numbers of edits to simulate a lifelong editing scenario.
From the single-edit perspective, our method exhibits optimal performance in most testing scenarios.
In the lifelong editing scenarios, we have the following observations:
(1) Methods that modify the parameters of LLMs show outstanding editing performance in a single edit. Yet, they exhibit a significant decline in editing performance as the number of edits increases. This trend aligns with the toxic accumulation issue highlighted by \citet{WILKE}.
(2) Method introducing additional parameters maintains a degree of reliability and generality in the lifelong editing process. However, the cumulative addition of extra parameters compromises the original inference process, evidenced by the pronounced deterioration in locality observed in ZSRE.
(3) Retrieval-based approaches demonstrate robustness against the increasing number of edits. Among them, our method achieves the best results, affirming the strengths of retrieval as well as validating the efficacy of our strategy. 

From the perspective of metrics, for reliability and generality, the co-occurrence of correct retrieval and the model's faithful adherence to editing instructions is a prerequisite for editing success. Therefore, compared to other retrieval-based methods, we consider the additional introduction of KS in RECIPE, which enhances retrieval performance and thereby improves editing efficacy, to be a crucial factor for RECIPE’s superior performance in terms of reliability and generality metrics, as demonstrated in Table 4. Regarding locality, aside from the ability of the editing retrieval to effectively exclude irrelevant knowledge, the locality component of the editing loss for RECIPE, along with the shorter continuous prompt editing prefixes, also minimizes the impact on the model's editing-unrelated responses.

\noindent\textbf{Damage to the General Performance of LLMs:}
While the three editing metrics effectively demonstrate the editing performance, we further investigate to which extent these editors influence the model's general capabilities.
Table \ref{tab_llm_gen_exp} shows the results of LLaMA-2 after 1,000 edits. It is observed that non-retrieval-based methods lead to a significant reduction in general capabilities.
This can be attributed to the accumulation of pattern mismatches caused by external interventions of editing.
Among retrieval-based methods, LTE also exhibits performance degradation. 
In contrast, our RECIPE does not involve direct intervention on LLM parameters but instead relies on concatenating a short prompt to guide the LLM’s adherence to knowledge.
It demonstrates the best preservation of general performance, suggesting that it inflicts minimal harm on the model.

\subsection{Efficiency Comparison}
To underscore the efficiency of RECIPE, we conduct a comparative analysis on editing and inference time after 10,000 edits, as delineated in Table~\ref{tab_time_compare}.
Among methods leveraging edit-specific training such as MEND, MALMEN, LTE, and RECIPE, a notable reduction in editing time is observed when compared to techniques necessitating multiple iterations of back-propagation during editing. 
For inference speed, methods that modify model parameters maintain consistent speeds as they do not alter the original inference pipeline. 
T-Patcher slows down the inference speed due to the accumulating neurons. Among retrieval-based methods, GRACE reduces the parallelism in model inference due to its unique dictionary pairing mechanism.
R-ROME and LTE need to calculate editing matrices on the fly and concatenate long editing instructions, respectively. 
In contrast, RECIPE effectively preserves the LLM's original inference speed by concatenating short continuous prompts for editing.
The shortest total time also highlights RECIPE's efficiency advantage.

\begin{figure}[!t]
    \centering
    \includegraphics[width=.95\linewidth]{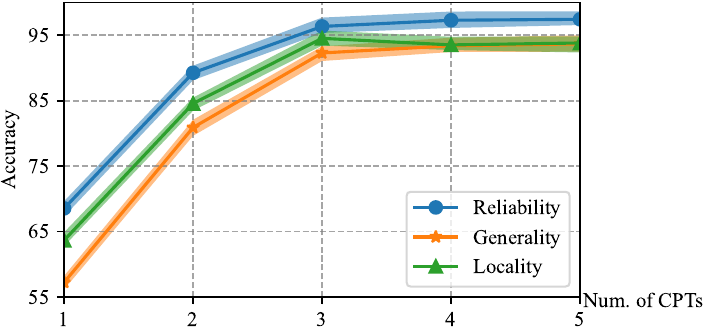}
    \caption{Impact of the number of CPTs on editing performance of RECIPE.}
    \label{fig_GCP_n_to_rgl}
\end{figure}

\begin{figure}[!t]
    \centering
    \includegraphics[width=.775\linewidth]{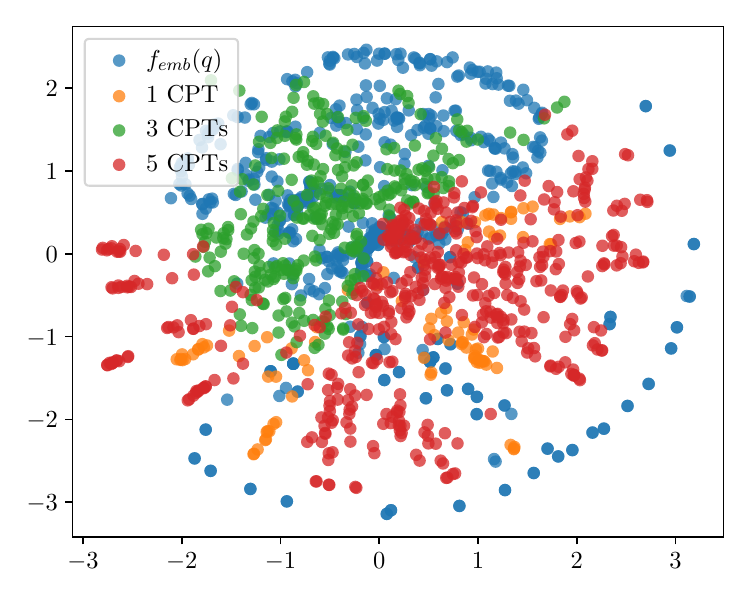}
    \caption{Visualization of word embeddings with varying numbers of CPTs.}
    \label{fig_GCP_dim_reduc_vis}
\end{figure}

\subsubsection{Number of Continuous Prompt Tokens} 
To assess whether an increase in Continuous Prompt Tokens (CPTs) can enhance the editing performance of RECIPE \cite{DBLP:journals/corr/abs-2001-08361}, Figure~\ref{fig_GCP_n_to_rgl} illustrates the average impact of varying CPTs on editing efficacy across the editing benchmarks after 1,000 edits. 
The results show a noticeable performance dip with a single CPT, particularly in generality, indicating that fewer tokens limit representational capacity and lead to learning overly-simple patterns. Optimal editing performance is observed with three CPTs. Beyond this, while reliability and generality improve modestly, locality slightly decreases. This suggests that more CPTs expand representational capabilities but also introduce additional LLMs' interference.

Regarding the peak editing performance with three CPTs, we suggest that this is because the information carried by edit facts can be succinctly represented as relational triples \texttt{(Head Entity, Relation, Tail Entity)}, and these triples can be represented as three word-level token embeddings.
Thus, we further visualize LLAMA-2's word embeddings of subjects and objects of 100 edit facts in CF, along with the corresponding representations of 1, 3, and 5 CPTs, reduced to two dimensions using t-SNE \cite{van2008visualizing}. 
From Figure \ref{fig_GCP_dim_reduc_vis}, the representations with three CPTs are closer to word embeddings than the others, indicating that the granularity of information carried by three CPTs is more akin to that of word embeddings of LLMs.

\subsection{Ablation Study}
We conduct an ablation study using LLAMA-2 on ZSRE~\cite{MEND}, CF \cite{ROME}, and RIPE \cite{RIPPLEDITS}. Average results are detailed in Table~\ref{tab_ablation}. Without CPTs, we resort to using word embeddings of knowledge statements as retrieval prompts from the knowledge repository. Excluding KS involved applying a conventional contrastive learning loss to align reliability and generality sample representations closer to editing knowledge while distancing those of locality samples.
Upon completion of training, we employ an absolute similarity threshold decision strategy~\cite{RASE} for filtering irrelevant knowledge. 
Despite its high locality, the omission of CPTs significantly impairs RECIPE's reliability and generality. It can be observed that the results are nearly identical to those obtained without using an editor at all.
This underscores that merely using raw concatenated knowledge prefixes fails to make LLMs comply with editing directives.
Conversely, CPTs aid LLM adherence to specified edits. 
Additionally, discarding KS leads to a deterioration in editing efficacy, particularly impacting generality and locality. The reason is that an absolute similarity threshold fails to adequately address the diverse thresholds required by distinct queries.

\begin{table}[!tb]
    \footnotesize
    \centering
    \setlength{\tabcolsep}{3.8pt}

\begin{tabular}{lcccccc}
\toprule
\multirow{2}{*}{\textbf{Settings}} & \multicolumn{3}{c}{\textbf{100 Edits}}           & \multicolumn{3}{c}{\textbf{1000 Edits}}          \\
                                   & \textbf{Rel.}  & \textbf{Gen.}  & \textbf{Loc.}  & \textbf{Rel.}  & \textbf{Gen.}  & \textbf{Loc.}  \\
\midrule
N/A                                & 27.30          & 26.07          & 100.00         & 27.30          & 26.07          & 100.00         \\
\midrule
RECIPE                             & \textbf{97.29} & \textbf{93.74} & 97.38          & \textbf{96.05} & \textbf{92.34} & 95.36          \\
- CPT                              & 27.42          & 26.18          & \textbf{99.98} & 27.38          & 26.15          & \textbf{99.97} \\
- KS                               & 95.55          & 89.10          & 92.45          & 94.01          & 86.63          & 88.55          \\
- BOTH                             & 27.41          & 26.17          & 99.96          & 27.35          & 26.12          & 99.94        \\ 
\bottomrule
\end{tabular}

\caption{Ablation study of RECIPE.}
    \label{tab_ablation}
\end{table}

\section{Conclusion}
We propose RECIPE, an effective and efficient LLM editor that includes two essential modules. 
Continuous prompt learning prefixes transformed knowledge to input query to achieve efficient post-retrieval editing.
Dynamic prompt retrieval with KS retrieves and determines whether the repository contains relevant knowledge without fixed similarity thresholds.
In lifelong editing, RECIPE demonstrates exceptional editing performance and efficiency while simultaneously preserving LLM functionality without degradation.

\section*{Limitations}
Due to the limitation in machine resources, we have not experimented on larger knowledge encoders apart from RoBERTa \cite{roberta} and larger LLMs. We speculate that either a larger encoder or a larger LLM may yield better editing performance. Additionally, the current editing experiments are primarily implemented on QA-based datasets. We will expand our RECIPE framework to other types of editing tasks and larger models in the future.

\section*{Acknowledgments}
This work is supported by the National Key Research and Development Program of China under Grant No. 2022ZD0120302.

\bibliography{custom}

\appendix

\begin{algorithm}
    \caption{Training of RECIPE}
    \small
    \newcommand{\comm}[1]{\textcolor{gray!50}{\textit{#1}}}
    \begin{algorithmic}[1]
        \STATE \textbf{Input:} LLM to be edited $f_{llm}$;
        initialized collection of RECIPE parameters $\mathcal{M}$;
        training set $\mathcal{D}=\left\{\left(
        (q_{e_i}, a_{e_i}), 
        \{q_{g_i}^{j}, a_{g_i}^{j}\}_{j=1}^{N_{g_i}},
        \{q_{l_i}^{j}, a_{l_i}^{j}\}_{j=1}^{N_{l_i}}
        \right)\right\}_{i=1}^N$;
        maximum iteration number $I_{max}$;
        batch size of training samples $b$; 
        learning rate $\eta$.
        \STATE \textbf{Output:} trained RECIPE parameters $\mathcal{M}$.
        \WHILE{$iter < I_{max}$}
            \STATE $\left\{(q_{e_i}, a_{e_i}), (q_{g_i}, a_{g_i}),(q_{l_i}, a_{l_i})
            \right\}_{i=1}^b\leftarrow$ Sample $b$ training samples from $\mathcal{D}$
            \FOR{$i \gets 1$ to $b$}
                \STATE \comm{\# Get editing knowledge}
                \STATE$k_i = q_{e_i}+a_{e_i}$ \comm{\# String concatenation}
                \STATE \comm{\# Get knowledge representation}
                \STATE$r_{k_i}  \leftarrow $ Transform $k_i$ using Eq. \ref{equation:knowledge_representation}
                \STATE \comm{\# Get continuous prompts}
                \STATE$p_{k_i} \leftarrow $ Transform $r_{k_i}$ using Eq. \ref{equation:prompt_embedding}
                \STATE \comm{\# Get query representations}
                \STATE$\tilde{r}_{q_{e_i}} \leftarrow $ Transform $q_{e_i}$ using Eq. \ref{equation:query_representation}
                \STATE$\tilde{r}_{q_{g_i}} \leftarrow $ Transform $q_{g_i}$ using Eq. \ref{equation:query_representation}
                \STATE$\tilde{r}_{q_{l_i}} \leftarrow $ Transform $q_{l_i}$ using Eq. \ref{equation:query_representation}
            \ENDFOR
            \STATE \comm{\# Get knowledge representation of KS}
            \STATE$r_{\Theta} \leftarrow$ Transform $\Theta$ using Eq. \ref{equation:knowledge_representation}
            \STATE \comm{\# Compute loss and update parameters}
            \STATE $\mathcal{L}_{edit},\mathcal{L}_{pl}\leftarrow$ Compute losses using Eq.\ref{eq_loss_edit} and Eq.\ref{eq_loss_pl} 
            \STATE $\mathcal{L}_{total} = \mathcal{L}_{edit}+\mathcal{L}_{pl}$
            \STATE$\mathcal{M}\leftarrow \operatorname{Adam}\left(\nabla_{\mathcal{M}} \mathcal{L}_{total}, \eta\right)$
        \ENDWHILE
        \RETURN $\mathcal{M}$
    \end{algorithmic}
    \label{alg_recipe_train}
\end{algorithm}

\section{Algorithms of RECIPE}
\label{sec_Algorithms_of_RECIPE}
The training and editing algorithms for RECIPE are detailed in Alg. \ref{alg_recipe_train} and Alg. \ref{alg_recipe_edit}, respectively. The inference process of the LLM equipped with RECIPE is described in Alg. \ref{alg_recipe_llm_infer}.

\begin{algorithm}[!tb]
    \caption{Editing of RECIPE in a Lifelong Scenario}
    \small
    \newcommand{\comm}[1]{\textcolor{gray!50}{\textit{#1}}}
    \begin{algorithmic}[1]
        \STATE \textbf{Input:} Knowledge retrieval repository $\mathcal{K}_{t-1}= \{(r_{k_\tau}, p_{k_\tau})\}_{\tau=1}^{t-1}$;
        editing knowledge $(q_{e_t}, a_{e_t})$.
        \STATE \textbf{Output:} updated knowledge retrieval repository $\mathcal{K}_t$.
            \STATE \comm{\# Get editing knowledge}
            \STATE$k_t = q_{e_t}+a_{e_t}$ \comm{\# String concatenation}
            \STATE \comm{\# Get knowledge representation}
            \STATE$r_{k_t} \leftarrow$ Transform $k_t$ using Eq. \ref{equation:knowledge_representation}
            \STATE \comm{\# Get continuous prompts}
            \STATE$p_{k_t} \leftarrow $ Transform $r_{k_t}$ using Eq. \ref{equation:prompt_embedding}
            \STATE \comm{\# Update knowledge retrieval repository}
            \STATE$\mathcal{K}_t = \mathcal{K}_{t-1} \cup \{(r_{k_t}, p_{k_t})\}$
        \RETURN $\mathcal{K}_t$
    \end{algorithmic}
    \label{alg_recipe_edit}
\end{algorithm}

\begin{algorithm}[!tb]
    \caption{Inference of LLM Equipped with RECIPE}
    \small
    \newcommand{\comm}[1]{\textcolor{gray!50}{\textit{#1}}}
    \begin{algorithmic}[1]
        \STATE \textbf{Input:} LLM to be edited $f_{llm}$,
        including the embedding layer $f_{emb}$ and the transformer module $\hat{f}_{llm}$;
        knowledge retrieval repository $\mathcal{K}_{t}= \{(r_{k_\tau}, p_{k_\tau})\}_{\tau=1}^{t}$; knowledge representation of KS $r_\Theta$;
        input query $q$.
        \STATE \textbf{Output:} LLM's output with RECIPE intervened $a_q$.
            \STATE \comm{\# Get query representation}
            \STATE$\tilde{r}_{q} = \mathbf{MLP}_Q(f_{rm}(q))$
            \STATE \comm{\# Get the index of knowledge with the largest similarity}
            \STATE  $j = \mathop{\arg\max}\limits_{\tau=1,\ldots,t}\; \tilde{r}_{q}^T \cdot r_{k_\tau}$
            \STATE \comm{\# Filter irrelevant knowledge and get output}
            \IF{$\tilde{r}_{q}^T \cdot r_{k_j}>\tilde{r}_{q}^T \cdot r_\Theta$}
                \STATE$a_{q} = \hat{f}_{llm}(p_{k_j} \oplus f_{emb}(q))$
            \ELSE
                \STATE$a_{q} = f_{llm}(q)$
            \ENDIF
        \RETURN $a_q$
    \end{algorithmic}
    \label{alg_recipe_llm_infer}
\end{algorithm}

\section{Datasets and Baselines}
\label{datasets_baselines}

\subsection{Model Editing Datasets}
We employ three public model editing datasets, including ZSRE \cite{MEND}, CounterFact (CF) \cite{ROME}, and Ripple Effect (RIPE) \cite{RIPPLEDITS} as our experimental datasets. 
For methods that require edit training, including MEND \cite{MEND}, 
MALMEN \cite{DBLP:journals/corr/abs-2311-04661}, LTE \cite{LTE}, and our RECIPE, we utilize the above training sets to learn their parameters.

\textbf{ZSRE} \citep{DBLP:conf/conll/LevySCZ17} is generated through question-answering with BART \citep{DBLP:conf/acl/LewisLGGMLSZ20} and manual filtering, including 162,555 training and 19,009 testing samples. Each sample comprises an editing sample and its rephrased and irrelevant counterparts, matching the reliability, generality, and locality editing properties. 

\textbf{CF} \citep{ROME} is characterized by the editing of false facts and includes 10,000 training and 10,000 testing samples. These false facts are more likely to conflict with the original knowledge within LLMs, making the editing process more challenging and thus providing a robust evaluation of the editors' ability to enforce edits. 

\textbf{RIPE} \citep{RIPPLEDITS} differentiates the generality and locality properties into fine-grained types, 
comprising 3,000 training and 1,388 testing samples. The generality of each sample includes logical generalization, combination I, combination II, and subject aliasing, while the locality data cover forgetfulness and relation specificity.

\subsection{General Datasets of LLMs}
To evaluate the damage of editors to the general performance of LLMs, 
we select four prevalent benchmarks to assess LLMs' general capabilities. They are 
CSQA \cite{DBLP:conf/naacl/TalmorHLB19} to evaluate commonsense knowledge,
ANLI \cite{DBLP:conf/acl/NieWDBWK20} for reasoning abilities,
MMLU \cite{DBLP:conf/iclr/HendrycksBBZMSS21} to gauge exam capabilities,
and SQuAD-2 \cite{DBLP:conf/acl/RajpurkarJL18} for comprehension skills.
PromptBench \cite{DBLP:journals/corr/abs-2312-07910} is utilized as the evaluation framework for this experiment.

\textbf{CSQA} (CommonSense Question Answering) \cite{DBLP:conf/naacl/TalmorHLB19} is designed to evaluate LLMs' commonsense knowledge through multiple-choice questions. 
It includes 12,102 samples, split into 9,741 for training, 1,221 for validation, and 1,140 for testing.

\textbf{ANLI} (Adversarial Natural Language Inference) \cite{DBLP:conf/acl/NieWDBWK20} 
evaluates LLMs' natural language reasoning capacity by determining whether the relationship between a premise and a hypothesis is one of entailment, contradiction, or neutrality.
The difficulty of the tasks increases across three rounds.
It includes a total of 169,265 samples, with 162,865 for training, 3,200 for validation, and 3,200 for testing.

\textbf{MMLU} (Massive Multitask Language Understanding) \cite{DBLP:conf/iclr/HendrycksBBZMSS21} 
tests LLMs' mastery of specialized domain knowledge through multiple-choice questions covering 57 different academic fields and disciplines, such as history, literature, law, and biology. The dataset comprises a total of 6,783 questions distributed across testing, validation, and development sets, containing 5,871, 627, and 285 samples, respectively.

\textbf{SQuAD-2} (Stanford Question Answering Dataset version 2) \cite{DBLP:conf/acl/RajpurkarJL18}  
assesses the reading comprehension abilities of LLMs by posing questions based on paragraphs taken from over 500 Wikipedia articles. Compared to its first version \cite{DBLP:conf/emnlp/RajpurkarZLL16}, its challenge lies in the inclusion of questions that do not have answers derivable from the text. The dataset contains a total of 142,192 questions, with 130,319 in the training set and 11,873 in the validation set. We report the performance on its validation set with default hyper-parameter settings.

Among them, we use selection accuracy as the evaluation metric for CSQA, MMLU, and ANLI, and the reciprocal of the model's perplexity (PPL) on the predicted sequence as the evaluation metric for SQuAD-2. All metrics are scaled to a range of 0-100.

\subsection{Baselines}
In addition to fine-tuning (FT) as the basic baseline, we compare our RECIPE approach with various strong editing baselines. 
\textbf{MEND} \citep{MEND} trains an MLP to transform the low-rank decomposition of the gradients of the model to be edited with respect to the editing samples.
\textbf{ROME} \citep{ROME} first uses causal mediation analysis to locate the layer that has the greatest impact on the editing sample. 
\textbf{MEMIT} \citep{MEMIT} expands the editing scope to multiple layers based on ROME, thereby improving editing performance and supporting batch editing. 
\textbf{T-Patcher} \citep{T-Patcher} (TP) attaches and trains additional neurons in the FFN of the last layer of the model to be edited. 
\textbf{MALMEN} \citep{DBLP:journals/corr/abs-2311-04661} formulates the parameter shift aggregation as a least square problem, subsequently updating the LM parameters using the normal equation.
\textbf{WILKE} \citep{WILKE} selects the editing layer based on the pattern matching degree of editing knowledge across different layers.
 
We also leverage competitive retrieval-based editing methods to validate the effectiveness further. 
\textbf{GRACE} \citep{GRACE} proposes retrieval adapters for continuous editing, which maintains a dictionary-like structure to construct new mappings for potential representations that need to be modified.
\textbf{RASE} \citep{RASE} leverages factual information to enhance editing generalization and guide the identification of edits by retrieving related facts from the fact-patch memory. In our baseline settings, we use the ROME \citep{ROME} model as the specific basic editor for RASE to perform the editing task, named \textbf{R-ROME}.  \textbf{LTE} \citep{LTE} elicits the capabilities of LLMs to follow knowledge editing instructions, thereby empowering them to effectively leverage updated knowledge to answer queries.

\section{Model Settings and Training Details}
\label{model_settings}
\noindent\textbf{RECIPE.}
(1) \textbf{Hyper-parameter Settings}: For our proposed RECIPE, we use the same hyper-parameter settings across different backbones, including LLAMA-2\footnote{\url{https://huggingface.co/meta-llama/Llama-2-7b-hf}}, GPT-J\footnote{\url{https://huggingface.co/EleutherAI/gpt-j-6b}}, and GPT2-XL\footnote{\url{https://huggingface.co/openai-community/gpt2-xl}}. The number of continuous prompt tokens and KS tokens are set as $l=3$ and $c=10$, respectively. $\mathbf{MLP}_K$, $\mathbf{MLP}_Q$, and $\mathbf{MLP}_P$ are each composed of two linear layers, with an intermediate dimension set to 4096 and are connected in a residual manner \cite{DBLP:conf/cvpr/HeZRS16}. The dimensions of the knowledge and query representations are also set to 4096. The total numbers of RECIPE's training parameters for GPT2-XL, GPT-J, and LLAMA-2 are 220M, 250M, and 250M, respectively.
(2) \textbf{Training Details}: 
We set the learning rate ($\eta=1e-5$), the batch size to 8, and the maximum number of iterations to 150,000. A checkpoint is saved every 5000 iterations, and ultimately, the one with the smallest loss is selected for evaluation. The training process requires approximately 3 days on an NVIDIA A800 GPU. These experiments are presented on average with $5$ random runs, using different random seeds but the same hyper-parameters.

\noindent\textbf{Baseline Models.}
For R-ROME \citep{RASE} and LTE \citep{LTE}, we implement the settings mentioned in their respective papers and trained them on the same datasets as ours. For the other baselines, we follow the same settings as described in EasyEdit \citep{DBLP:journals/corr/abs-2308-07269} for training and evaluation.

\section{Results with Different Backbones}
\label{result_diff_llm}
Lifelong editing experiments on GPT-J (6B) and GPT2-XL (1.5B) are 
are presented in Table \ref{tab_main_exp_gpt_j} and Table \ref{tab_main_exp_gpt2_xl}. 
The results also show a similar conclusion with general results, demonstrating the efficacy of our method.
Notably, comparing Tables \ref{tab_main_exp}, \ref{tab_main_exp_gpt_j}, and \ref{tab_main_exp_gpt2_xl}, ROME and MEMIT exhibit a significant performance decline on LLAMA-2 compared to the other two backbones. MEMIT is an improved version of ROME, and both methods aim to edit the weights of the selected FFN layers. We speculate that this performance discrepancy may stem from structural differences in the FFN layers between LLAMA-2, GPT2-XL, and GPT-J. The original ROME and MEMIT papers assume a common two-layer structure for the LLM's FFN layers, which is exactly the configuration of GPT2-XL and GPT-J. However, the FFN of LLAMA-2 consists of three linear layers, which may account for the ineffectiveness of ROME and MEMIT in adapting to the LLAMA-2 architecture.

\begin{table*}[!tb]
    \scriptsize
    \centering
    \setlength{\tabcolsep}{3.8pt}
    \renewcommand{\arraystretch}{0.85}
    \begin{tabular}{cccccccccccccccc}

\toprule
 &\multirow{2}{*}{\textbf{\# Editing}} & \multirow{2}{*}{\textbf{Type}} & \multirow{2}{*}{\textbf{Editor}} & \multicolumn{4}{c}{\textbf{ZSRE}}                                          & \multicolumn{4}{c}{\textbf{CF}}                                            & \multicolumn{4}{c}{\textbf{RIPE}}                                          \\
 &       &                     &                         & \textbf{Rel.}           & \textbf{Gen.}           & \textbf{Loc.}           & \textbf{Avg.}           & \textbf{Rel.}           & \textbf{Gen.}           & \textbf{Loc.}           & \textbf{Avg.}           & \textbf{Rel.}           & \textbf{Gen.}           & \textbf{Loc.}           & \textbf{Avg.}           \\ 
\midrule
&\multirow{11}{*}{1}& \multirow{6}{*}{MP}&FT &80.22 &84.58 &45.51 &70.11$_{(\pm1.43)}$ &98.11 &42.10 &42.10 &60.77$_{(\pm0.93)}$ &75.14 &51.12 &15.94 &47.40$_{(\pm0.57)}$ \\
&& &MEND &54.43 &59.17 &90.21 &67.93$_{(\pm0.80)}$ &72.59 &70.19 &91.26 &78.01$_{(\pm1.44)}$ &31.52 &10.03 &19.13 &20.22$_{(\pm0.29)}$ \\
&& &ROME &99.14 &95.76 &99.53 &98.14$_{(\pm0.44)}$ &\textbf{99.62} &83.61 &95.87 &93.04$_{(\pm0.36)}$ &\textbf{99.42} &39.55 &39.71 &59.56$_{(\pm1.14)}$ \\
&& &MEMIT &99.64 &86.83 &99.51 &95.33$_{(\pm1.18)}$ &99.13 &38.98 &95.69 &77.93$_{(\pm0.69)}$ &99.14 &33.60 &51.14 &61.29$_{(\pm0.58)}$ \\
&& &MALMEN &59.29 &58.59 &6.34 &41.41$_{(\pm1.02)}$ &22.94 &21.28 &15.00 &19.74$_{(\pm0.62)}$ &59.05 &36.26 &13.95 &36.42$_{(\pm0.70)}$ \\
&& &WILKE &97.95 &94.40 &97.65 &96.67$_{(\pm0.85)}$ &97.82 &82.97 &94.42 &91.73$_{(\pm1.18)}$ &98.27 &41.13 &39.07 &59.49$_{(\pm0.55)}$ \\
\cmidrule(lr){3-16}
&& \multirow{1}{*}{AP}&TP &94.66 &93.27 &90.92 &92.95$_{(\pm0.93)}$ &99.33 &61.03 &13.86 &58.07$_{(\pm0.30)}$ &90.91 &60.46 &36.40 &62.59$_{(\pm0.52)}$ \\
\cmidrule(lr){3-16}
&& \multirow{4}{*}{RB}&GRACE &99.29 &14.20 &99.49 &71.00$_{(\pm0.81)}$ &99.59 &0.01 &98.14 &65.91$_{(\pm1.25)}$ &99.12 &21.95 &99.50 &73.52$_{(\pm1.67)}$ \\
&& &R-ROME &96.75 &92.33 &98.62 &95.90$_{(\pm0.67)}$ &96.57 &80.64 &97.77 &91.66$_{(\pm0.42)}$ &95.86 &35.51 &92.63 &74.66$_{(\pm1.12)}$ \\
&& &LTE &98.98 &98.58 &98.81 &98.79$_{(\pm0.18)}$ &98.88 &98.10 &91.95 &96.31$_{(\pm0.86)}$ &98.90 &84.87 &87.42 &90.40$_{(\pm0.74)}$ \\
&& & RECIPE &\textbf{99.70} &\textbf{99.42} &\textbf{99.98} &\textbf{99.70}$_{(\pm0.04)}$ &98.72 &\textbf{98.55} &\textbf{98.67} &\textbf{98.65}$_{(\pm0.44)}$ &98.95 &\textbf{85.51} &\textbf{99.60} &\textbf{94.69}$_{(\pm1.06)}$ \\
\midrule
&\multirow{11}{*}{10}& \multirow{6}{*}{MP}&FT &30.14 &23.04 &3.14 &18.77$_{(\pm0.96)}$ &96.09 &35.67 &23.89 &51.88$_{(\pm0.40)}$ &29.87 &17.81 &4.06 &17.24$_{(\pm0.36)}$ \\
&& &MEND &0.37 &0.41 &0.56 &0.44$_{(\pm0.22)}$ &0.59 &0.17 &0.19 &0.31$_{(\pm0.14)}$ &0.00 &0.03 &0.04 &0.02$_{(\pm0.01)}$ \\
&& &ROME &81.06 &78.75 &94.62 &84.81$_{(\pm0.95)}$ &95.94 &59.41 &90.02 &81.79$_{(\pm0.15)}$ &98.18 &41.84 &39.15 &59.72$_{(\pm0.43)}$ \\
&& &MEMIT &82.04 &75.99 &94.68 &84.23$_{(\pm0.67)}$ &96.02 &38.03 &95.46 &76.50$_{(\pm1.14)}$ &98.52 &37.73 &47.31 &61.19$_{(\pm0.75)}$ \\
&& &MALMEN &98.86 &98.35 &92.00 &96.41$_{(\pm0.84)}$ &90.02 &32.86 &77.11 &66.67$_{(\pm0.93)}$ &89.72 &68.08 &57.62 &71.81$_{(\pm0.80)}$ \\
&& &WILKE &84.09 &82.71 &95.82 &87.54$_{(\pm0.41)}$ &96.97 &68.00 &92.72 &85.90$_{(\pm0.99)}$ &94.52 &40.32 &35.24 &56.69$_{(\pm0.67)}$ \\
\cmidrule(lr){3-16}
&& \multirow{1}{*}{AP}&TP &85.20 &78.29 &77.19 &80.23$_{(\pm1.29)}$ &96.02 &54.31 &3.61 &51.31$_{(\pm0.53)}$ &80.83 &56.72 &32.39 &56.64$_{(\pm0.49)}$ \\
\cmidrule(lr){3-16}
&& \multirow{4}{*}{RB}&GRACE &48.08 &21.74 &98.88 &56.23$_{(\pm0.30)}$ &66.50 &0.89 &96.43 &54.61$_{(\pm1.35)}$ &45.15 &21.06 &97.16 &54.45$_{(\pm0.55)}$ \\
&& &R-ROME &94.40 &86.48 &98.09 &92.99$_{(\pm0.52)}$ &94.71 &76.09 &95.76 &88.85$_{(\pm0.95)}$ &94.90 &32.56 &84.95 &70.80$_{(\pm0.54)}$ \\
&& &LTE &98.34 &97.53 &98.34 &98.07$_{(\pm0.90)}$ &97.55 &97.19 &91.26 &95.34$_{(\pm0.35)}$ &97.85 &84.26 &86.82 &89.65$_{(\pm0.85)}$ \\
&& &RECIPE &\textbf{98.91} &\textbf{98.71} &\textbf{99.98} &\textbf{99.20}$_{(\pm0.49)}$ &\textbf{97.88} &\textbf{97.63} &\textbf{97.38} &\textbf{97.63}$_{(\pm0.63)}$ &\textbf{98.58} &\textbf{84.95} &\textbf{99.00} &\textbf{94.18}$_{(\pm1.15)}$ \\
\midrule
&\multirow{11}{*}{100}& \multirow{6}{*}{MP}&FT &20.37 &10.04 &0.70 &10.37$_{(\pm0.41)}$ &66.70 &15.69 &2.66 &28.35$_{(\pm0.34)}$ &16.49 &8.50 &2.40 &9.13$_{(\pm0.93)}$ \\
&& &MEND &0.18 &0.13 &0.01 &0.11$_{(\pm0.02)}$ &0.13 &0.15 &0.02 &0.10$_{(\pm0.03)}$ &0.02 &0.01 &0.09 &0.04$_{(\pm0.02)}$ \\
&& &ROME &77.44 &75.59 &84.99 &79.34$_{(\pm0.96)}$ &78.79 &38.43 &52.13 &56.45$_{(\pm0.81)}$ &95.69 &35.93 &32.15 &54.59$_{(\pm1.12)}$ \\
&& &MEMIT &77.95 &74.10 &90.22 &80.76$_{(\pm0.32)}$ &94.09 &40.24 &85.15 &73.16$_{(\pm0.98)}$ &86.61 &33.32 &33.46 &51.13$_{(\pm0.51)}$ \\
&& &MALMEN &50.58 &40.74 &59.25 &50.19$_{(\pm1.16)}$ &29.64 &31.78 &67.99 &43.13$_{(\pm0.30)}$ &39.93 &27.78 &53.26 &40.32$_{(\pm0.43)}$ \\
&& &WILKE &80.41 &78.67 &86.68 &81.92$_{(\pm0.76)}$ &81.90 &48.33 &64.03 &64.75$_{(\pm0.33)}$ &91.63 &36.43 &32.85 &53.63$_{(\pm0.18)}$ \\
\cmidrule(lr){3-16}
&& \multirow{1}{*}{AP}&TP &68.52 &59.31 &52.77 &60.20$_{(\pm0.66)}$ &75.99 &31.90 &2.25 &36.71$_{(\pm0.56)}$ &64.22 &36.42 &23.65 &41.43$_{(\pm0.87)}$ \\
\cmidrule(lr){3-16}
&& \multirow{4}{*}{RB}&GRACE &46.27 &21.00 &98.05 &55.11$_{(\pm0.41)}$ &52.34 &0.69 &93.70 &48.91$_{(\pm0.94)}$ &42.75 &20.90 &94.26 &52.64$_{(\pm0.63)}$ \\
&& &R-ROME &94.37 &78.08 &96.95 &89.80$_{(\pm0.45)}$ &90.64 &69.60 &93.46 &84.56$_{(\pm0.33)}$ &92.62 &28.49 &77.36 &66.15$_{(\pm0.51)}$ \\
&& &LTE &97.17 &97.03 &98.95 &97.72$_{(\pm1.05)}$ &96.28 &96.05 &90.68 &94.34$_{(\pm0.44)}$ &97.17 &83.46 &82.29 &87.64$_{(\pm0.61)}$ \\
&& &RECIPE &\textbf{98.83} &\textbf{98.15} &\textbf{99.97} &\textbf{98.98}$_{(\pm0.63)}$ &\textbf{96.87} &\textbf{96.37} &\textbf{96.31} &\textbf{96.52}$_{(\pm0.61)}$ &\textbf{97.64} &\textbf{84.36} &\textbf{95.48} &\textbf{92.49}$_{(\pm0.27)}$ \\
\midrule
&\multirow{11}{*}{1000}& \multirow{6}{*}{MP}&FT &12.61 &7.78 &0.19 &6.86$_{(\pm0.68)}$ &31.59 &8.21 &1.41 &13.74$_{(\pm0.23)}$ &9.06 &3.09 &1.20 &4.45$_{(\pm1.62)}$ \\
&& &MEND &0.01 &0.01 &0.03 &0.02$_{(\pm0.01)}$ &0.02 &0.01 &0.06 &0.03$_{(\pm0.00)}$ &0.16 &0.13 &0.08 &0.12$_{(\pm0.02)}$ \\
&& &ROME &57.19 &53.89 &29.88 &46.98$_{(\pm0.84)}$ &0.17 &0.25 &0.62 &0.35$_{(\pm0.08)}$ &47.50 &16.97 &13.40 &25.96$_{(\pm0.54)}$ \\
&& &MEMIT &56.83 &54.56 &54.90 &55.43$_{(\pm0.79)}$ &82.36 &36.41 &30.64 &49.80$_{(\pm0.54)}$ &0.01 &0.00 &0.02 &0.01$_{(\pm0.00)}$ \\
&& &MALMEN &43.00 &35.09 &39.26 &39.12$_{(\pm0.49)}$ &15.06 &12.36 &25.06 &17.49$_{(\pm1.56)}$ &31.06 &19.10 &35.33 &28.50$_{(\pm0.86)}$ \\
&& &WILKE &69.35 &67.63 &48.78 &61.92$_{(\pm0.73)}$ &15.66 &12.85 &29.06 &19.19$_{(\pm0.66)}$ &64.25 &30.70 &25.07 &40.01$_{(\pm1.10)}$ \\
\cmidrule(lr){3-16}
&& \multirow{1}{*}{AP}&TP &45.71 &40.39 &10.53 &32.21$_{(\pm0.87)}$ &47.33 &17.02 &1.47 &21.94$_{(\pm0.51)}$ &48.09 &29.08 &15.18 &30.78$_{(\pm0.13)}$ \\
\cmidrule(lr){3-16}
&& \multirow{4}{*}{RB}&GRACE &47.70 &20.40 &97.15 &55.08$_{(\pm0.73)}$ &46.36 &0.50 &90.18 &45.68$_{(\pm0.37)}$ &39.89 &20.58 &88.20 &49.56$_{(\pm0.96)}$ \\
&& &R-ROME &91.63 &68.72 &94.78 &85.04$_{(\pm0.29)}$ &88.83 &56.26 &89.94 &78.34$_{(\pm0.92)}$ &85.83 &24.74 &67.53 &59.37$_{(\pm0.19)}$ \\
&& &LTE &96.67 &96.27 &99.11 &97.35$_{(\pm0.76)}$ &94.76 &93.16 &88.37 &92.10$_{(\pm0.75)}$ &94.82 &81.31 &74.67 &83.60$_{(\pm0.89)}$ \\
&& &RECIPE &\textbf{97.45} &\textbf{96.71} &\textbf{99.96} &\textbf{98.05}$_{(\pm0.54)}$ &\textbf{95.82} &\textbf{95.40} &\textbf{92.04} &\textbf{94.42}$_{(\pm0.71)}$ &\textbf{95.28} &\textbf{83.55} &\textbf{89.16} &\textbf{89.33}$_{(\pm0.99)}$ \\
\midrule
&\multirow{11}{*}{10000}& \multirow{6}{*}{MP}&FT &8.37 &4.54 &0.10 &4.34$_{(\pm0.51)}$ &21.89 &9.57 &1.53 &11.00$_{(\pm1.09)}$ &- &- &- &- \\
&& &MEND &0.02 &0.01 &0.03 &0.02$_{(\pm0.01)}$ &0.01 &0.01 &0.03 &0.02$_{(\pm0.00)}$ &- &- &- &- \\
&& &ROME &12.49 &10.84 &3.16 &8.83$_{(\pm0.61)}$ &0.12 &0.16 &0.48 &0.25$_{(\pm0.12)}$ &- &- &- &- \\
&& &MEMIT &0.01 &0.01 &0.02 &0.01$_{(\pm0.00)}$ &0.01 &0.03 &0.03 &0.02$_{(\pm0.01)}$ &- &- &- &- \\
&& &MALMEN &25.15 &12.45 &21.83 &19.81$_{(\pm0.96)}$ &8.20 &4.10 &10.86 &7.72$_{(\pm1.15)}$ &- &- &- &- \\
&& &WILKE &27.04 &20.70 &10.23 &19.32$_{(\pm0.79)}$ &5.09 &2.09 &12.78 &6.65$_{(\pm0.86)}$ &- &- &- &- \\
\cmidrule(lr){3-16}
&& \multirow{1}{*}{AP}&TP &31.40 &27.36 &3.79 &20.85$_{(\pm0.83)}$ &25.95 &8.96 &0.98 &11.96$_{(\pm1.57)}$ &- &- &- &- \\
\cmidrule(lr){3-16}
&& \multirow{4}{*}{RB}&GRACE &45.76 &21.33 &95.17 &54.08$_{(\pm1.34)}$ &42.40 &0.77 &85.39 &42.85$_{(\pm0.25)}$ &- &- &- &- \\
&& &R-ROME &84.99 &50.07 &93.01 &76.02$_{(\pm0.67)}$ &82.28 &42.79 &86.42 &70.49$_{(\pm0.22)}$ &- &- &- &- \\
&& &LTE &94.29 &89.03 &99.19 &94.17$_{(\pm1.16)}$ &92.84 &91.47 &82.24 &88.85$_{(\pm1.37)}$ &- &- &- &- \\
&& &RECIPE &\textbf{94.79} &\textbf{89.85} &\textbf{99.92} &\textbf{94.85}$_{(\pm0.18)}$ &\textbf{93.37} &\textbf{92.01} &\textbf{89.53} &\textbf{91.63}$_{(\pm0.29)}$ &- &- &- &- \\
\bottomrule

    \end{tabular} 
    
    \caption{The overall results using GPT-J (6B) in lifelong editing scenario. ``\# Editing'' denotes the number of edits. ``Rel.'', ``Gen.'' and ``Loc.'' are the abbreviations of reliability, generality, and locality, respectively. Given that the RIPE dataset comprises 4,388 samples, achieving results for 10,000 edits is not feasible. MP, AP, and RB indicate Modifying Parameters, Adding Parameters, and Retrieval-Based methods, respectively. The t-tests demonstrate the improvements of our work are statistically significant with $p$ < 0.05 level.}
    
    \label{tab_main_exp_gpt_j}
\end{table*}

\begin{table*}[!tb]
    \scriptsize
    \centering
    \setlength{\tabcolsep}{3.8pt}
    \renewcommand{\arraystretch}{0.85}
    \begin{tabular}{cccccccccccccccc}

\toprule
 &\multirow{2}{*}{\textbf{\# Editing}} & \multirow{2}{*}{\textbf{Type}} & \multirow{2}{*}{\textbf{Editor}} & \multicolumn{4}{c}{\textbf{ZSRE}}                                          & \multicolumn{4}{c}{\textbf{CF}}                                            & \multicolumn{4}{c}{\textbf{RIPE}}                                          \\
 &         &                   &                         & \textbf{Rel.}           & \textbf{Gen.}           & \textbf{Loc.}           & \textbf{Avg.}           & \textbf{Rel.}           & \textbf{Gen.}           & \textbf{Loc.}           & \textbf{Avg.}           & \textbf{Rel.}           & \textbf{Gen.}           & \textbf{Loc.}           & \textbf{Avg.}           \\ 
\midrule
&\multirow{11}{*}{1}& \multirow{6}{*}{MP}&FT &81.00 &78.41 &79.78 &79.73$_{(\pm0.66)}$ &96.11 &33.99 &52.84 &60.98$_{(\pm0.60)}$ &63.59 &44.60 &38.28 &48.82$_{(\pm1.39)}$ \\
&& &MEND &87.24 &80.14 &76.51 &81.30$_{(\pm0.72)}$ &91.15 &83.94 &75.16 &83.41$_{(\pm0.97)}$ &52.16 &26.03 &30.23 &36.14$_{(\pm1.03)}$ \\
&& &ROME &99.50 &86.91 &99.29 &95.23$_{(\pm0.72)}$ &\textbf{99.37} &45.41 &95.99 &80.26$_{(\pm0.71)}$ &98.93 &40.73 &42.39 &60.68$_{(\pm0.75)}$ \\
&& &MEMIT &69.11 &50.05 &99.29 &72.82$_{(\pm0.31)}$ &79.02 &26.42 &98.22 &67.89$_{(\pm0.82)}$ &59.09 &28.69 &60.07 &49.29$_{(\pm1.24)}$ \\
&& &MALMEN &54.72 &66.14 &14.99 &45.28$_{(\pm0.69)}$ &48.08 &24.27 &6.14 &26.16$_{(\pm1.35)}$ &62.81 &28.53 &10.16 &33.83$_{(\pm1.16)}$ \\
&& &WILKE &97.94 &85.09 &97.92 &93.65$_{(\pm0.51)}$ &98.02 &43.14 &94.60 &78.58$_{(\pm0.22)}$ &97.82 &45.85 &51.16 &64.94$_{(\pm0.98)}$ \\
\cmidrule(lr){3-16}
&& \multirow{1}{*}{AP}&TP &39.41 &36.82 &97.49 &57.91$_{(\pm1.56)}$ &38.36 &6.83 &89.04 &44.74$_{(\pm1.26)}$ &54.98 &36.21 &87.56 &59.58$_{(\pm0.51)}$ \\
\cmidrule(lr){3-16}
&& \multirow{4}{*}{RB}&GRACE &99.58 &16.75 &99.32 &71.88$_{(\pm1.30)}$ &99.31 &0.02 &\textbf{99.13} &66.16$_{(\pm1.01)}$ &\textbf{99.33} &17.85 &99.15 &72.11$_{(\pm0.43)}$ \\
&& &R-ROME &98.22 &85.28 &99.20 &94.23$_{(\pm0.20)}$ &98.10 &44.62 &98.56 &80.43$_{(\pm0.83)}$ &97.36 &38.66 &93.62 &76.54$_{(\pm0.89)}$ \\
&& &LTE &99.09 &98.70 &98.36 &98.72$_{(\pm0.44)}$ &98.59 &98.14 &89.05 &95.26$_{(\pm0.37)}$ &98.81 &77.88 &76.24 &84.31$_{(\pm0.34)}$ \\
&& &RECIPE &\textbf{99.68} &\textbf{99.01} &\textbf{99.99} &\textbf{99.56}$_{(\pm0.11)}$ &98.74 &\textbf{98.38} &98.68 &\textbf{98.60}$_{(\pm0.65)}$ &98.79 &\textbf{78.52} &\textbf{99.32} &\textbf{92.21}$_{(\pm1.16)}$ \\
\midrule
&\multirow{11}{*}{10}& \multirow{6}{*}{MP}&FT &54.87 &49.03 &46.99 &50.30$_{(\pm1.21)}$ &91.48 &28.44 &23.09 &47.67$_{(\pm0.67)}$ &36.73 &21.65 &22.82 &27.07$_{(\pm0.95)}$ \\
&& &MEND &7.57 &6.67 &7.55 &7.26$_{(\pm0.28)}$ &8.48 &3.83 &3.06 &5.12$_{(\pm0.65)}$ &9.79 &5.26 &6.16 &7.07$_{(\pm0.27)}$ \\
&& &ROME &81.05 &76.80 &96.74 &84.87$_{(\pm0.27)}$ &96.63 &44.00 &89.40 &76.68$_{(\pm0.97)}$ &97.90 &40.19 &34.53 &57.54$_{(\pm0.11)}$ \\
&& &MEMIT &70.87 &60.37 &98.55 &76.60$_{(\pm0.28)}$ &80.21 &27.01 &96.63 &67.95$_{(\pm0.83)}$ &62.78 &30.56 &52.47 &48.60$_{(\pm0.32)}$ \\
&& &MALMEN &92.10 &88.88 &90.27 &90.42$_{(\pm1.08)}$ &86.55 &32.09 &63.43 &60.69$_{(\pm1.09)}$ &78.45 &54.96 &76.81 &70.07$_{(\pm0.21)}$ \\
&& &WILKE &76.06 &73.26 &94.74 &81.35$_{(\pm0.28)}$ &92.77 &40.34 &83.20 &72.11$_{(\pm0.99)}$ &93.27 &43.63 &50.70 &62.54$_{(\pm0.21)}$ \\
\cmidrule(lr){3-16}
&& \multirow{1}{*}{AP}&TP &47.57 &41.18 &90.34 &59.70$_{(\pm0.74)}$ &43.85 &8.45 &60.77 &37.69$_{(\pm0.99)}$ &59.21 &37.04 &68.26 &54.84$_{(\pm0.66)}$ \\
\cmidrule(lr){3-16}
&& \multirow{4}{*}{RB}&GRACE &47.26 &17.70 &98.71 &54.56$_{(\pm0.33)}$ &66.90 &0.02 &97.40 &54.78$_{(\pm0.72)}$ &37.69 &19.24 &97.41 &51.45$_{(\pm1.26)}$ \\
&& &R-ROME &96.88 &81.43 &99.56 &92.62$_{(\pm0.84)}$ &95.71 &41.26 &97.57 &78.18$_{(\pm1.22)}$ &96.58 &37.36 &87.08 &73.67$_{(\pm0.45)}$ \\
&& &LTE &98.49 &98.01 &96.60 &97.70$_{(\pm0.88)}$ &98.05 &97.60 &87.13 &94.26$_{(\pm1.62)}$ &98.15 &74.21 &74.94 &82.43$_{(\pm1.36)}$ \\
&& &RECIPE &\textbf{98.82} &\textbf{98.59} &\textbf{99.98} &\textbf{99.13}$_{(\pm0.18)}$ &\textbf{98.23} &\textbf{97.72} &\textbf{97.89} &\textbf{97.95}$_{(\pm0.68)}$ &\textbf{98.58} &\textbf{75.13} &\textbf{97.64} &\textbf{90.45}$_{(\pm0.96)}$ \\
\midrule
&\multirow{11}{*}{100}& \multirow{6}{*}{MP}&FT &36.45 &31.92 &8.83 &25.73$_{(\pm0.58)}$ &40.67 &8.99 &3.67 &17.78$_{(\pm0.68)}$ &8.90 &4.19 &3.51 &5.53$_{(\pm0.27)}$ \\
&& &MEND &0.02 &0.02 &0.01 &0.01$_{(\pm0.01)}$ &0.01 &0.03 &0.01 &0.02$_{(\pm0.00)}$ &0.01 &0.02 &0.00 &0.01$_{(\pm0.00)}$ \\
&& &ROME &75.29 &70.75 &82.87 &76.31$_{(\pm0.79)}$ &63.70 &35.60 &37.89 &45.73$_{(\pm0.75)}$ &94.80 &43.68 &29.02 &55.83$_{(\pm0.88)}$ \\
&& &MEMIT &71.07 &63.60 &92.90 &75.86$_{(\pm0.89)}$ &86.52 &30.68 &86.30 &67.83$_{(\pm0.41)}$ &72.91 &34.27 &44.38 &50.52$_{(\pm0.43)}$ \\
&& &MALMEN &57.12 &49.45 &44.50 &50.36$_{(\pm0.75)}$ &33.75 &30.35 &58.16 &40.75$_{(\pm0.43)}$ &45.49 &39.68 &59.47 &48.21$_{(\pm1.25)}$ \\
&& &WILKE &71.49 &69.30 &85.78 &75.52$_{(\pm0.80)}$ &72.72 &36.33 &49.36 &52.80$_{(\pm1.10)}$ &80.56 &37.26 &36.47 &51.43$_{(\pm0.59)}$ \\
\cmidrule(lr){3-16}
&& \multirow{1}{*}{AP}&TP &51.98 &46.46 &78.48 &58.97$_{(\pm0.37)}$ &42.00 &8.02 &14.66 &21.56$_{(\pm0.85)}$ &54.11 &38.08 &47.76 &46.65$_{(\pm1.47)}$ \\
\cmidrule(lr){3-16}
&& \multirow{4}{*}{RB}&GRACE &43.38 &19.24 &95.81 &52.81$_{(\pm0.69)}$ &63.24 &0.68 &95.82 &53.25$_{(\pm1.59)}$ &33.06 &18.54 &94.28 &48.63$_{(\pm0.63)}$ \\
&& &R-ROME &95.94 &74.45 &98.76 &89.72$_{(\pm0.49)}$ &91.75 &37.59 &96.11 &75.15$_{(\pm0.31)}$ &93.47 &35.32 &80.23 &69.67$_{(\pm1.47)}$ \\
&& &LTE &96.77 &96.06 &94.72 &95.85$_{(\pm0.57)}$ &97.14 &97.07 &84.35 &92.85$_{(\pm1.07)}$ &96.16 &67.19 &72.47 &78.61$_{(\pm0.50)}$ \\
&& &RECIPE &\textbf{98.67} &\textbf{98.56} &\textbf{99.98} &\textbf{99.07}$_{(\pm0.31)}$ &\textbf{97.22} &\textbf{97.10} &\textbf{96.19} &\textbf{96.84}$_{(\pm0.41)}$ &\textbf{97.32} &\textbf{70.42} &\textbf{94.32} &\textbf{87.35}$_{(\pm0.26)}$ \\
\midrule
&\multirow{11}{*}{1000}& \multirow{6}{*}{MP}&FT &25.61 &18.52 &1.23 &15.12$_{(\pm1.87)}$ &28.69 &8.72 &2.41 &13.27$_{(\pm0.85)}$ &4.72 &1.67 &0.66 &2.35$_{(\pm0.71)}$ \\
&& &MEND &0.07 &0.05 &1.85 &0.66$_{(\pm0.23)}$ &0.01 &0.02 &0.02 &0.02$_{(\pm0.01)}$ &0.02 &0.01 &0.00 &0.01$_{(\pm0.00)}$ \\
&& &ROME &44.54 &37.47 &43.09 &41.70$_{(\pm0.26)}$ &0.82 &0.89 &1.01 &0.91$_{(\pm0.19)}$ &43.72 &16.06 &17.08 &25.62$_{(\pm0.34)}$ \\
&& &MEMIT &57.31 &50.85 &48.21 &52.12$_{(\pm0.82)}$ &80.72 &48.13 &24.14 &51.00$_{(\pm0.36)}$ &28.82 &15.72 &21.59 &22.04$_{(\pm0.20)}$ \\
&& &MALMEN &29.32 &35.44 &35.05 &33.27$_{(\pm0.26)}$ &12.37 &13.73 &34.03 &20.04$_{(\pm0.79)}$ &21.84 &23.76 &31.99 &25.86$_{(\pm1.02)}$ \\
&& &WILKE &48.13 &43.87 &55.51 &49.17$_{(\pm0.44)}$ &46.29 &22.68 &19.70 &29.55$_{(\pm0.54)}$ &55.10 &25.48 &33.49 &38.02$_{(\pm0.47)}$ \\
\cmidrule(lr){3-16}
&& \multirow{1}{*}{AP}&TP &45.97 &42.68 &60.46 &49.70$_{(\pm0.67)}$ &27.78 &7.20 &5.72 &13.57$_{(\pm0.68)}$ &47.71 &33.24 &31.04 &37.33$_{(\pm0.78)}$ \\
\cmidrule(lr){3-16}
&& \multirow{4}{*}{RB}&GRACE &48.86 &19.73 &93.75 &54.11$_{(\pm0.24)}$ &63.83 &0.52 &92.52 &52.29$_{(\pm0.97)}$ &33.18 &19.80 &\textbf{90.81} &47.93$_{(\pm0.79)}$ \\
&& &R-ROME &94.48 &67.99 &98.87 &87.11$_{(\pm1.08)}$ &89.01 &31.51 &92.86 &71.13$_{(\pm0.60)}$ &88.87 &33.15 &72.19 &64.73$_{(\pm0.79)}$ \\
&& &LTE &94.73 &92.27 &91.10 &92.70$_{(\pm0.56)}$ &95.13 &94.31 &81.28 &90.24$_{(\pm1.05)}$ &92.18 &62.78 &66.51 &73.82$_{(\pm0.98)}$ \\
&& &RECIPE &\textbf{96.94} &\textbf{96.43} &\textbf{99.98} &\textbf{97.79}$_{(\pm0.31)}$ &\textbf{96.86} &\textbf{96.33} &\textbf{93.70} &\textbf{95.63}$_{(\pm0.32)}$ &\textbf{94.25} &\textbf{67.78} &89.85 &\textbf{83.96}$_{(\pm0.54)}$ \\
\midrule
&\multirow{11}{*}{10000}& \multirow{6}{*}{MP}&FT &15.54 &11.94 &1.96 &9.81$_{(\pm0.73)}$ &21.91 &7.92 &2.04 &10.62$_{(\pm1.18)}$ &- &- &- &- \\
&& &MEND &0.06 &0.09 &1.85 &0.67$_{(\pm0.16)}$ &0.01 &0.00 &0.01 &0.01$_{(\pm0.00)}$ &- &- &- &- \\
&& &ROME &17.79 &14.19 &1.23 &11.07$_{(\pm0.84)}$ &0.30 &0.41 &0.07 &0.26$_{(\pm0.04)}$ &- &- &- &- \\
&& &MEMIT &0.02 &0.00 &0.01 &0.01$_{(\pm0.01)}$ &0.25 &0.25 &0.05 &0.18$_{(\pm0.06)}$ &- &- &- &- \\
&& &MALMEN &7.81 &11.13 &4.97 &7.97$_{(\pm1.01)}$ &6.06 &4.22 &18.22 &9.50$_{(\pm0.33)}$ &- &- &- &- \\
&& &WILKE &26.94 &23.62 &11.86 &20.81$_{(\pm0.59)}$ &27.03 &14.91 &15.13 &19.02$_{(\pm1.16)}$ &- &- &- &- \\
\cmidrule(lr){3-16}
&& \multirow{1}{*}{AP}&TP &36.60 &34.79 &17.51 &29.63$_{(\pm1.03)}$ &19.70 &9.11 &2.75 &10.52$_{(\pm1.20)}$ &- &- &- &- \\
\cmidrule(lr){3-16}
&& \multirow{4}{*}{RB}&GRACE &49.81 &20.45 &91.48 &53.91$_{(\pm1.49)}$ &64.19 &0.48 &87.28 &50.65$_{(\pm0.19)}$ &- &- &- &- \\
&& &R-ROME &89.17 &54.69 &97.48 &80.44$_{(\pm0.42)}$ &84.14 &23.59 &87.01 &64.91$_{(\pm1.26)}$ &- &- &- &- \\
&& &LTE &89.85 &87.17 &88.66 &88.56$_{(\pm0.47)}$ &92.38 &89.17 &76.82 &86.13$_{(\pm0.59)}$ &- &- &- &- \\
&& &RECIPE &\textbf{90.61} &\textbf{89.29} &\textbf{99.99} &\textbf{93.29}$_{(\pm0.57)}$ &\textbf{93.72} &\textbf{92.73} &\textbf{88.49} &\textbf{91.65}$_{(\pm1.33)}$ &- &- &- &- \\
\bottomrule

    \end{tabular} 
    
    \caption{The overall results using GPT2-XL (1.5B) in lifelong editing scenario. ``\# Editing'' denotes the number of edits. ``Rel.'', ``Gen.'' and ``Loc.'' are the abbreviations of reliability, generality, and locality, respectively. Given that the RIPE dataset comprises 4,388 samples, achieving results for 10,000 edits is not feasible. MP, AP, and RB indicate Modifying Parameters, Adding Parameters, and Retrieval-Based methods, respectively. The t-tests demonstrate the improvements of our work are statistically significant with $p$ < 0.05 level.}
    
    \label{tab_main_exp_gpt2_xl}
\end{table*}

\section{Summary of Innovations}
We summarize the innovations of this work as follows.

First, to the best of our knowledge, RECIPE is the first method to employ prompt learning for solving model editing. Indeed, previous works have used string-constructed prefixes to accomplish model editing, such as IKE \cite{InContextKnowledgeEdit} and LTE \cite{LTE}. However, we are the first to explore how to achieve model editing with the shortest editing prefixes through prompt learning. 
Second, a direct application of prompt learning might involve training a continuous prompt as a fixed prefix, which can append any editing string behind it to form a complete instruction for the LLM to follow. This approach treats the entire editing task as a prompt learning task. However, the concatenation of the continuous prompt with the editing string results in a lengthy prefix. Moreover, it can be reasonably hypothesized that a smaller amount of trainable parameters might make it challenging to converge to satisfactory editing performance. In contrast, our RECIPE innovatively treats the editing of each piece of knowledge as a mini prompt learning task, and considers the training of the encoder that generates continuous editing prompts as a meta-task. This strategy significantly reduces the length of editing prefixes while enhancing the representational capabilities of the editing training, thus ensuring editing performance.
Third, RECIPE completely decouples knowledge editing from model parameters through prompt learning, thereby avoiding model degradation due to external intervention in model parameters or overfitting on the editing dataset.

\end{document}